\documentclass[10pt,twocolumn,letterpaper]{article}

\usepackage{cvpr}              % To produce the CAMERA-READY version % Enable 2:00pm 28 Mar 2026
\definecolor{cvprblue}{rgb}{0.21,0.49,0.74}
\usepackage[pagebackref,breaklinks,colorlinks,allcolors=cvprblue]{hyperref}
\usepackage{tjchinmacros}
\usepackage{multirow}
\usepackage{algorithm}
\usepackage{amsmath,amsfonts}
\usepackage{algpseudocode}
\usepackage[accsupp]{axessibility}  % Improves PDF readability for those with disabilities.
\algnewcommand{\Inputs}[1]{%
  \State \textbf{Inputs:}
  \Statex \hspace*{\algorithmicindent}\parbox[t]{.99\linewidth}{\raggedright #1}
}
\algnewcommand{\Outputs}[1]{%
  \State \textbf{Outputs:}
  \Statex \hspace*{\algorithmicindent}\parbox[t]{.99\linewidth}{\raggedright #1}
}
\algnewcommand{\Initialize}[1]{%
  \State \textbf{Initialize:}
  \Statex \hspace*{\algorithmicindent}\parbox[t]{.8\linewidth}{\raggedright #1}
}
\algnewcommand{\LeftComment}[1]{\Statex \(\triangleright\) #1}

\def\paperID{} % *** Enter the Paper ID here
\def\confName{CVPR}
\def\confYear{2026}

\title{Uncertainty-Guided Edge Learning for Deep Image Regression in Remote Sensing}

\author{Anh Vu Nguyen\and Dino Sejdinovic\and Tat-Jun Chin\\
Australian Institute for Machine Learning (AIML), Adelaide University\\
{\tt\small \{anhvu.nguyen, dino.sejdinovic, tat-jun.chin\}@adelaide.edu.au}
}

\begin{document}
\maketitle
\begin{abstract}
% Edge learning refers to training machine learning models deployed on edge platforms, typically using new data accumulated onboard. The computational limitations on edge devices affect not only model optimisation, but also calculation of the predictive uncertainty of the current model on the unlabelled data, which is vital for informing model updating. In this paper, we investigate edge learning in the context of performing deep image regression on a remote sensing satellite, where a deep network is executed by an onboard computer to regress a scalar $y$ from an input image, \emph{e.g.}, $y$ is the percentage of pixels indicating cloud coverage or land use. We propose an uncertainty-guided edge learning (UGEL) algorithm that can accurately prioritise the data to speed up training convergence of the onboard regression model. Underpinning UGEL is the calculation of predictive uncertainty based on deep beta regression, where a deep network is used to estimate the parameters of a beta distribution for which the target $y$ for an input image has a high likelihood. Compared to established methods for uncertainty estimation that are either too costly on edge devices (\textit{e.g.}, require many forward passes per sample) or make strict assumptions on the predictive distribution (\textit{e.g.}, Gaussian), deep beta regression is computable in a single forward pass and allows more general predictive distributions. Results show that UGEL delivers faster-converging edge learning than active or semi-supervised learning\footnote{Source code will be released after the review period.}.

Edge learning refers to training machine learning models deployed on edge platforms, typically using new data accumulated onboard. The computational limitations on edge devices affect not only model optimisation, but also calculation of the predictive uncertainty of the current model on the unlabelled data, which is vital for informing model updating. In this paper, we investigate edge learning in the context of performing deep image regression on a remote sensing satellite, where a deep network is executed by an onboard computer to regress a scalar $y$ from an input image, \emph{e.g.}, $y$ is the percentage of pixels indicating cloud coverage or land use. We propose an uncertainty-guided edge learning (UGEL) algorithm that can accurately prioritise the data to speed up training convergence of the onboard regression model. Underpinning UGEL is the calculation of predictive uncertainty based on deep beta regression, where a deep network is used to estimate the parameters of a beta distribution for which the target $y$ for an input image has a high likelihood. Compared to established methods for uncertainty estimation that are either too costly on edge devices (\textit{e.g.}, require many forward passes per sample) or make strict assumptions on the predictive distribution (\textit{e.g.}, Gaussian), deep beta regression is computable in a single forward pass and allows more general predictive distributions. Results show that UGEL delivers faster-converging edge learning than active or semi-supervised learning. Code and models are publicly available at \url{https://github.com/anh-vunguyen/UGEL}.
\end{abstract}    
\section{Introduction}
\label{sec:intro}
Deploying machine learning models on the edge promises the benefits of lower latency operation, reduced data transmission costs, and enhanced privacy~\cite{chen2019deep, wang2020convergence, martinez2021deep, cai2022enable}. However, a deployed model invariably needs to be updated to maintain its accuracy. Edge learning (EL) aims to train a machine learning model on the edge~\cite{khouas2024training, zhu2024device, dhar2021survey}. A major theme is performing the update using data accumulated on the platform itself, since such data reflects the actual operating environment.

\begin{figure}[t]
    \centering
    \includegraphics[width=0.99\columnwidth]{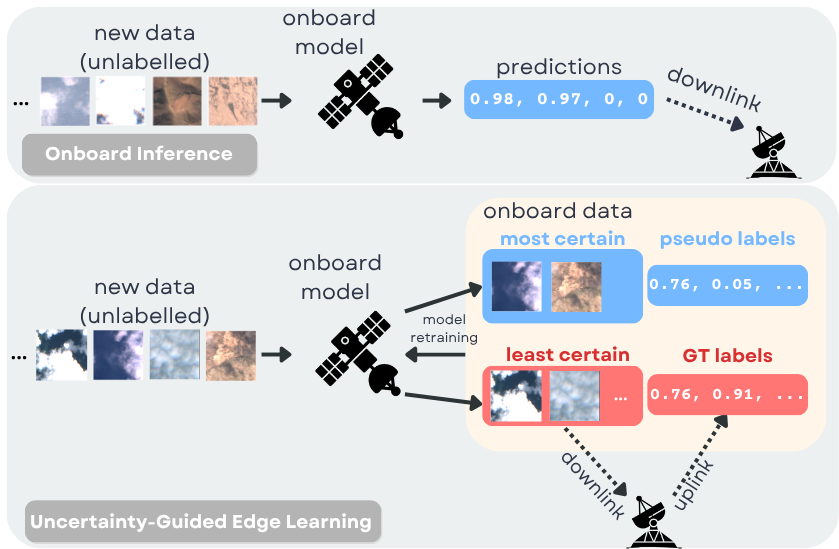}
    \caption{(Top) Onboard inference for responsive RS missions. (Bottom) The proposed UGEL for RS missions.}
    \label{fig:scenario}
\end{figure}

Progress in space-hardening neural network accelerators for use on orbit has opened up the potential of satellite-borne machine learning~\cite{Felix2024total,Rodriguez-Ferrandez2024proton}. In particular, in remote sensing (RS) missions, by performing inference on Earth imagery collected by onboard imagers and using the prediction for onboard decision making, the time and energy costs of downlinking the data are avoided, thus enabling more responsive missions; see Fig.~\ref{fig:scenario}. To motivate using examples, let $f_{\hat{\bw}}$ be a deep network with weights $\hat{\bw}$, and $\hat{y} = f_{\hat{\bw}}(x)$ be the onboard evaluation of $f_{\hat{\bw}}$ on an Earth image $x$. If $\hat{y}$ is the percentage of pixels that are cloud-occluded, $x$ can be discarded if $\hat{y}$ is high, thus saving onboard storage~\cite{giuffrida2020cloudscout}. If $\hat{y}$ is the percentage of pixels indicating land cover, timely survey of environmental impact can be obtained without needing to downlink $x$ for further processing~\cite{finer2018combating}.

The need for EL on RS missions arises from the domain gap between the testing data $\cI = \{ x_i \}^{N}_{i=1}$ collected onboard and the dataset used to train $f_{\hat{\bw}}$ due to inherent sensor discrepancies and drift~\cite{du2023domain}. Thus, $f_{\hat{\bw}}$ needs to be periodically updated to ensure accuracy in long-term operation. Despite the increasing potential for satellite-borne neural network accelerators, the compute capability will likely be much lower than that of terrestrial versions. Moreover, satellite platforms have stringent power constraints. This demands the update on $f_{\hat{\bw}}$ to be computational- and energy-efficient, which  invariably translates into using only a small subset $\cT \subset \cI$ of the unlabelled data to retrain $f_{\hat{\bw}}$.

Choosing the subset $\cT$ for model retraining and extracting supervisory signals from the unlabelled data are core problems in EL~\cite{khouas2024training}. These tasks are related to active learning (AL)~\cite{ren2021survey}, which optimises data subset selection for human labelling, and semi-supervised learning (SSL)~\cite{yang2022survey}, which leverages pre-trained models to assign labels to new data. Underpinning the selection of $\cT$ is the estimation of the predictive uncertainty of $f_{\hat{\bw}}$ on $\cI$~\cite{mcclarren2018uncertainty}. However, if $f_{\hat{\bw}}$ has not adapted to the operating environment, balancing the contributions of AL and SSL to selecting $\cT$ is difficult.

Moreover, since $|\cI| \gg |\cT|$, the cost of uncertainty estimation can actually dominate that of retraining $f_{\hat{\bw}}$ on $\cT$, and an expensive uncertainty estimation technique can exacerbate the computational challenges to EL. For example, the established Monte Carlo (MC) dropout method~\cite{gal2016dropout} requires multiple forward passes (\emph{e.g.}, 10 to 30) per sample to collect posterior statistics, which can quickly drain the energy of the edge platform.

By adopting a non-Bayesian approach and placing evidential priors over the Gaussian likelihood function, the seminal deep evidential regression (DER) method~\cite{amini2020deep} directly predicts the hyperparameters of the evidential distribution and avoids multiple forward passes during inference. However, Gaussian predictive uncertainties are unsuitable for some applications, \emph{e.g.}, in the RS tasks of cloud and forest coverage prediction, $y$ is bounded in $[0,1]$ whereas a Gaussian distribution is unbounded. Applying DER runs the risk of biased uncertainties, and since the accuracy of uncertainty estimation directly influences the usefulness of the selected $\cT$ for updating $f_{\hat{\bw}}$, the convergence of EL could be impacted. In short, existing uncertainty estimation methods are unsuitable for the task at hand.

% We investigate EL in the context of deep image regression for RS applications. Our key contributions are:
To summarise, our key contributions are:
\begin{itemize}
    \item We propose uncertainty-guided edge learning (UGEL) that unites AL and SSL in a principled manner through a common uncertainty estimation step. We will show clear evidence that combining human labels and pseudo-labels on small subsets of $\cT$ chosen in UGEL allows the updating of $f_{\hat{\bw}}$ to converge much more rapidly than AL, SSL, or a simple combination of AL and SSL~\cite{rottmann2018deep}.   
    \item We propose the usage of deep beta regression (DBR), where a deep network predicts the beta distribution for which the target $y$ for an input image $x$ has a high likelihood, as the uncertainty estimator for UGEL. This permits more flexible predictive distributions and respects the target bound $y \in [0,1]$. Also crucially, the differential entropy of beta distribution is a theoretically justified uncertainty estimate which can be computed in one forward pass, which is attractive for edge application.
    \item We demonstrate that the proposed framework consistently outperforms strong baselines across multiple image regression tasks with various modern lightweight backbones through extensive and rigorous experiments.
\end{itemize}
% Fig.~\ref{fig:scenario} illustrates UGEL with DBR. Note that while the beta regression model is well-known (see, \emph{e.g.}, \citet{cribari2010beta}), using DBR in the context of EL is new.
\section{Related works}
\label{sec:related_works}
The need to update machine learning models deployed on the edge motivates EL~\cite{khouas2024training,zhu2024device, dhar2021survey}. In addition to computational efficiency, major considerations in EL include model simplification, collaborative learning, and learning from partially labelled or unlabelled data---we focus on the latter in this paper. Techniques for EL from unlabelled data include unsupervised learning, self-supervised learning, transfer learning, AL and SSL, the latter two being the most relevant for our work. In the following, we will survey the relevant developments, including the core topic of uncertainty estimation.

%-------------------------------------------------------------------------
\subsection{Active learning}
The general aim of AL~\cite{li2024survey, ren2021survey, tharwat2023survey, settles2009active} is selecting a small subset of $\cI$ for human (groundtruth) annotation, which is then used to retrain $f_{\hat{\bw}}$. Finding the most appropriate subset under the manual labelling and data transfer budget (in settings where the subset needs to be transferred off board for human labelling) is a key AL problem. Broadly speaking, there are three strategies for AL: uncertainty-based~\cite{scheffer2001active, gal2017deep, yan2019label}, where samples with low prediction confidence by the current model are prioritised, diversity-based~\cite{sener2017active, tan2021diversity}, where samples that are representative of the underlying data distribution are selected, and hybrid techniques~\cite{ash2019deep, kirsch2019batchbald},  that combine the above approaches.

While clear benefits to non-deep learning methods have been recorded~\cite{settles2009active}, the utility of AL on deep learning is less obvious~\cite{munjal2022towards}. It has been conjectured that the small number of samples selected, which is generally of the same size as training batches, dilutes the effectiveness of AL for deep learning~\cite{tuia2024artificial}.

%-------------------------------------------------------------------------
\subsection{Semi-supervised learning}

SSL focusses on settings where we have a limited number of labelled (groundtruth) samples but a large number of potentially useful unlabelled data~\cite{yang2022survey}. Broadly the SSL loss combines supervised loss, unsupervised loss and relevant regularisation terms. Major paradigms include generative methods~\cite{springenberg2015unsupervised,denton2016semi,dai2017good}, where the data distribution is approximated before the joint data-label distribution is evaluated on the new samples, graph-based methods~\cite{abu2019mixhop,hamilton2017inductive,zhou2020towards}, where a neighbourhood structure is estimated for label propagation, consistency regularisation methods~\cite{berthelot2019mixmatch,berthelot2019remixmatch,sohn2020fixmatch}, which assume model predictions remain unchanged under realistic perturbations to data or model weights, and pseudo-labelling methods~\cite{lee2013pseudo,qiao2018deep,dong2018tri}, which generate labels for the new data based on one or more models trained on the labelled subset.

In EL settings where there is domain gap or shift, SSL should not rely on groundtruth samples acquired prior to deployment, since this will lead to bias in the decisions on the new unlabelled samples. We argue that the groundtruth samples should be extracted from the unlabelled samples directly, specifically using AL to request manual annotations on the appropriate subset of samples.

%-------------------------------------------------------------------------
\subsection{Active semi-supervised learning}

Active semi-supervised learning (ASSL) combines AL and SSL to alleviate their respective shortcomings. However, previous works on ASSL \cite{ma2025integrating,chen2023semi} assume all data to be located on a monolithic platform (\emph{e.g.}, a data centre) without restrictions on the computational and communication budget inherent to EL. More fundamentally, previous ASSL works combine AL and SSL simplistically, whereby AL is used to procure the small set of groundtruth samples for SSL.

We argue that more principled ways to conduct ASSL could lead to further benefits, \emph{e.g.}, cross leveraging the intermediate results of AL and SSL to improve accuracy and/or achieve computational savings. To this end, Sec.~\ref{sec:assledge} will describe our proposed UGEL algorithm.

%-------------------------------------------------------------------------
\subsection{Uncertainty estimation}

As mentioned above, uncertainty estimation is a key step in AL. The main approaches to uncertainty estimation are Bayesian inference, ensemble methods, and deterministic methods~\cite{wang2025uncertainty}. Due to the resource constraints of edge devices, ensemble methods---which require storing multiple models, performing several inferences per input, and updating all models in the ensemble when necessary---are not suitable and hence will not be considered.

Given training data $\cK = \{ (x_k, y_k) \}^{K}_{k=1}$, Bayesian inference defines the predictive distribution on new sample $x$ as
% \begin{align}\label{equ:bayesian}
%     p(y \mid x, \cK) = \int_\bw p(f_\bw(x) \mid x, \bw) p( \bw \mid \cK ) d\bw,
% \end{align}
\begin{align}\label{equ:bayesian}
    p(y \mid x, \cK) = \int_\bw p(y \mid x, \bw) p( \bw \mid \cK ) d\bw,
\end{align}
where $p(\bw \mid \cK)$ is the posterior of model weights $\bw$, and $p(y \mid x, \bw)$ is the likelihood. MC dropout \cite{gal2016dropout} approximates \eqref{equ:bayesian}  efficiently by mimicking a sample from the posterior distribution: first, a deep network $f_{\hat{\bw}}$ is trained on $\cK$ using dropout, where $\hat{\bw}$ are the learned weights. Then, $P$ forward passes of $x$ through $f_{\hat{\bw}}$ are conducted, with stochastic dropouts on $\hat{\bw}$ in each pass, yielding $P$ estimates $\{ \hat{y}^{(p)} \}^{P}_{p=1}$. The mean and variance of~\eqref{equ:bayesian} are then obtained respectively as
\begin{align}\label{eq:mcdropout}
    \hat{y} = \frac{1}{P}\sum^{P}_{p=1} \hat{y}^{(p)}, \;\;\;\; \hat{\sigma}^2 = \frac{1}{P} \sum^{P}_{p=1} (\hat{y}^{(p)} - \hat{y})^2,
\end{align}
with $\hat{y}$ and $\hat{\sigma}$ interpreted as the prediction at $x$ and its uncertainty. Since $P$ evaluations of $f_{\hat{\bw}}$ are required, MC dropout can be costly on edge devices if $P$ is large, \emph{e.g.}, $P = 30$ as suggested by \citet{gal2016dropout}.

DER~\cite{amini2020deep}, which is representative of deterministic methods, jointly predicts the regression target and its corresponding evidence to estimate predictive uncertainty. Under the assumption that regression targets follows a Gaussian distribution of unknown mean $\mu \sim \mathcal{N}(\gamma, \sigma^{2}\upsilon^{-1})$ and unknown variance $\sigma \sim \Gamma^{-1}(\alpha, \beta)$, DER optimises the model to infer the parameters $(\gamma, \upsilon, \alpha, \beta)$ of the evidential distribution given input $x$, which are then used to estimate the aleatoric and epistemic uncertainties
\begin{align}
   h_{a} = \mathop{\mathbb{E}}[\sigma^{2}] = \frac{\beta}{\alpha - 1}, \;\;\;\;  h_{e} = Var[\mu] = \frac{\beta}{\upsilon(\alpha - 1)}.
\end{align}
The sum $h_{a} + h_{e}$ is used as the predictive uncertainty on $x$ \cite{abdar2021review}. As we will show later, using DER in selecting $\cT$ in our UGEL method did not lead to consistently better performance than random sampling.

\section{Uncertainty-guided edge learning}\label{sec:assledge}

Let $f_{\hat{\bw}}$ be the deep regression model deployed onboard that was pre-trained with an \emph{initial} labelled dataset $\cD  = \{ (x_j, y_j) \}^{M}_{j=1}$, where $y_j \in [0,1]$. Details on training $f_{\hat{\bw}}$ will be given in Sec.~\ref{sec:betareg}. Our goal is updating $f_{\hat{\bw}}$ using the unlabelled data $\cI = \{ x_i \}^{N}_{i=1}$ collected onboard, where $N \gg M$.

The proposed UGEL algorithm for updating $f_{\hat{\bw}}$ using $\cI$ iteratively selects a subset $\cT \subset \cI$, labels the subset, and updates $f_{\hat{\bw}}$ across a number of rounds---a single iteration of UGEL is defined in Alg.~\ref{alg:ugel}. To facilitate labelling, a \emph{twin model} $f_{\mathring{\bw}}$ is also maintained on the edge. Both $f_{\hat{\bw}}$ and $f_{\mathring{\bw}}$ have the same architecture and were pre-trained on the same data $\cD$, but initialised differently, thus $\hat{\bw} \ne \mathring{\bw}$. Details of the main steps in Alg.~\ref{alg:ugel} are given in the following.

\subsection{Subset selection}

Let $h_{\hat{\bw}}(x_i)$ be the estimated measure of predictive uncertainty of $f_{\hat{\bw}}$ on $x_i$. Since UGEL is agnostic to the uncertainty estimation method, we will leave the description of the proposed DBR method to Sec.~\ref{sec:betareg}.

In each round of UGEL, $\cT = \cT_U \cup \cT_C$ is composed from
\begin{itemize}%[topsep=0em,leftmargin=1em,itemsep=0em]
\item A subset $\cT_U \subset \cI$ of size $B_U$ on which $f_{\hat{\bw}}$ is the \emph{most uncertain}, \emph{i.e.}, 
\begin{equation}\label{eq:uncertainty_al}
  \cT_{U} = \underset{\substack{\cS \subset\cI,~|\cS| = B_U}}{\arg\max} \sum_{x_i \in \cS} h_{\hat{\bw}}(x_i).
\end{equation}
\item A subset $\cT_C \subset \cI$ of size $B_C$ on which $f_{\hat{\bw}}$ is the \emph{most certain}, \emph{i.e.}, 
\begin{equation}\label{eq:uncertainty_ssl}
  \cT_{C} = \underset{\substack{\cS \subset\cI,~|\cS| = B_C}}{\arg\min} \sum_{x_i \in \cS} h_{\hat{\bw}}(x_i).
\end{equation}
\end{itemize}
By construction, $\cT_{U} \cap \cT_{C} = \emptyset$, hence $|\cT| = B_U + B_C$.

\begin{algorithm}[ht]
\caption{UGEL (single round)}
\label{alg:ugel}
\begin{algorithmic}[1]
\Inputs{
$f_{\hat{\bw}}$: regression model\\
$h_{\hat{\bw}}$: uncertainty estimator\\
$f_{\mathring{\bw}}$: twin model\\
$\cD = \{ (x_j, y_j) \}^{M}_{j=1} := (\cX, \cY)$: init.~labelled dataset\\
$\cI = \{ x_i \}^{N}_{i=1}$: unlabelled data (typ.~$N \gg M$)\\
$B_{U}$: \#samples for human annot.~(req.~$B_U \le N$)\\ 
$B_{C}$: \#samples for model annot.~(req.~$B_C \le N - B_U$)\\
$\tau$: loss coefficient
}
\LeftComment{\textbf{Select data subset}}
\State{$\cT_{U} \gets \underset{\substack{\cZ \subset\cI,~|\cZ| = B_U}}{\arg\max} \sum_{x_i \in \cZ} h_{\hat{\bw}}(x_i)$} 
\State{$\cT_{C} \gets \underset{\substack{\cZ \subset\cI,~|\cZ| = B_C}}{\arg\min} \sum_{x_i \in \cZ} h_{\hat{\bw}}(x_i)$}
\LeftComment{\textbf{Label data subset}}
\State{$\cY_{U} \gets \textrm{manually annotate}~\cT_U$}
\State{$\hat{\cY}_{C} \gets f_{\hat{\bw}}(\cT_{C})$ \hspace{3em} // Pseudo-labels}
\State{$\mathring{\cY}_{C} \gets f_{\mathring{\bw}}(\cT_{C})$ \hspace{3em} // Pseudo-labels}
\LeftComment{\textbf{Retrain model(s)}}
\State{$(\cX,\cY) \gets (\cX,\cY)  \cup  (\cT_{U},\cY_{U})$}
\State{$\hat{\cY} \gets f_{\hat{\bw}}(\cX)$ \hspace{4em} // Prediction}
\State{$\mathring{\cY} \gets f_{\mathring{\bw}}(\cX)$ \hspace{4em} // Prediction}
\State{$L\gets\cL_{s}(\hat{\cY} , \cY)+\cL_{s}(\mathring{\cY}, \cY) + \tau \cL_{RMSE}(\hat{\cY}\cup\hat{\cY}_{C}, \mathring{\cY}\cup\mathring{\cY}_{C})$}
\State{$f_{\hat{\bw}} \gets \textrm{model\_update}(f_{\hat{\bw}}, L)$}
\State{$f_{\mathring{\bw}} \gets \textrm{model\_update}(f_{\mathring{\bw}}, L)$}
\Outputs{
$f_{\hat{\bw}}$: updated regression model\\
$f_{\mathring{\bw}}$: updated twin model\\
$\cD \gets (\cX, \cY)$: expanded labelled dataset\\
$\cI \gets \cI  \setminus  \cT_{U}$: reduced unlabelled data}
\end{algorithmic}
\end{algorithm}
\vspace{-15pt}
\subsection{Label assignment}

Labelling of $\cT$ is achieved as follows:
\begin{itemize}%[topsep=0em,leftmargin=1em,itemsep=0em]
\item $\cT_U$ is transferred off board for manual annotation, yielding the label set $\cY_U$. Therefore, size $B_U$ should reflect the annotation budget, which depends on the available human effort and communication bandwidth.

\item $\cT_C$ is labelled by evaluating $f_{\hat{\bw}}$ and $f_{\mathring{\bw}}$ to yield the label sets $\hat{\cY}_C$ and $\mathring{\cY}_C$. Since $\cT_C$ has no groundtruth labels, $\hat{\cY}_C$ and $\mathring{\cY}_C$ play the role of \emph{pseudo-labels}.

\end{itemize}
Note that in the context of satellite-borne machine learning, $\cT_U$ needs to be downlinked with the labels $\cY_U$ subsequently uplinked. Hence $B_U$ should be small (more details below).

\subsection{Model retraining}

The loss function $L$ is a weighted combination of supervised loss $\cL_s$ and semi-supervised loss $\cL_{RMSE}$ (precise definitions of the losses will be given in Sec.~\ref{sec:betareg}). $\cL_{s}$ depends on the human-labelled samples only (including $(\cT_U,\cY_U)$) while $\cL_{RMSE}$ measures the inconsistency of the prediction of $f_{\hat{\bw}}$ and $f_{\mathring{\bw}}$ on the data considered thus far (including the pseudo-labelled samples). Intuitively, minimising $\cL_{RMSE}$ induces cross supervision~\cite{chen2021semi} between the primary and twin models.

The overall loss $L$ is then used to update both $f_{\hat{\bw}}$ and $f_{\mathring{\bw}}$. Then, $(\cT_U,\cY_U)$ is appended to $\cD$, $\cT_U$ is removed from $\cI$, and UGEL can proceed to the next round if $|\cI| \ge B_U$.

\subsection{Intuition behind UGEL}

UGEL allows to balance the contributions of human- and pseudo-labelled data, both chosen carefully using an uncertainty estimator, to speed up the training convergence. Executing UGEL sequentially will progressively expand $\cD$ and reduce $\cI$. However, since $\cD$ is initially small (\emph{e.g.}, $M = 12$ in typical online learning settings) and $\cT_U$ is kept small to minimise human labelling effort and data transfer load (\emph{e.g.}, $B_U = 6$ in our experiments), $f_{\hat{\bw}}$, $f_{\mathring{\bw}}$ and $h_{\hat{\bw}}$ tend to be inaccurate and uninformative initially. Using all pseudo-labelled samples from $\cI$, which dwarfs $\cD$ (\emph{e.g.}, 
size $N$ of $\cI$ can be $10k$), will lead to divergence in training. The size of $\cT_C$ should thus reflect the size of $\cD$. To this end, we use $B_C = |\cD|$, \emph{i.e.}, $\cT_C$ is small initially and grows with $\cD$.

Following from the above, we can also see that UGEL bridges AL, SSL and ASSL through the parameter settings:
\begin{itemize}
    \item If $B_U > 0$ and $B_C = 0$, UGEL reduces to uncertainty-based AL~\cite{scheffer2001active, gal2017deep}.
    \item If $B_U = 0$ and $B_C = N$, UGEL can be seen as SSL with cross pseudo supervision~\cite{chen2021semi} (in this setting, UGEL will terminate after one round).
    \item If $B_U > 0$ and $B_C = N - B_U$, UGEL reduces to the ASSL algorithm of~\citet{yao2023cloud} that uses all human- and pseudo-labelled samples in the training.
\end{itemize}
Sec.~\ref{sec:results} will show that being able to balance the relative contributions of human- and pseudo-labelled samples allows UGEL to converge much faster than AL, SSL and ASSL.

\subsection{Computational considerations}

In each round of UGEL, the uncertainty estimator $h_{\hat{\bw}}$ needs to be applied on the unlabelled set $\cI$ of size $N$. The samples involved in retraining $f_{\hat{\bw}}$ and $f_{\mathring{\bw}}$ are $\cD$, $\cT_U$ and $\cT_C$, which are of sizes $M$, $B_U$ and $B_C$ respectively. Since $N$ is much larger than $M$ and $B_U$, the total cost of uncertainty estimations in each round dominates that of retraining the models. This argues for a computationally efficient $h_{\hat{\bw}}$.
\section{Deep beta regression}\label{sec:betareg}

We focus on regressing a scalar $y \in [0,1]$ from an input image $x$ due its relevance to RS applications (see Sec.~\ref{sec:intro}). 

Our method DBR trains a deep network $f_{\bw}$ that outputs the parameters of a beta distribution given $x$. The probability density function of a beta distribution has a support on $[0,1]$ and two shape parameters $\alpha > 0$ and $\beta > 0$, \emph{i.e.},
\begin{align}
    p(y \mid \alpha, \beta) = \frac{1}{Beta(\alpha,\beta)} y^{\alpha-1} (1 - y)^{\beta-1}, \;\;\;\; y \in [0,1],
\end{align}
where $Beta(\alpha,\beta)$ is the beta function. We use an alternative parameterisation via the mean $\mu = \alpha/(\alpha + \beta)\in(0,1)$ and the precision $\nu = \alpha + \beta>0$, leading to
\begin{equation}
  p(y|\mu, \nu) = \frac{\Gamma(\nu)}{\Gamma(\mu\nu)\Gamma\bigl((1-\mu)\nu\bigl)}y^{\mu\nu-1}(1-y)^{(1-\mu)\nu-1},
  \label{eq:likelihood}
\end{equation}
where $\Gamma(\cdot)$ is the gamma function. Hence the output range of $f_{\bw}$ is $(0,1) \times \mathbb{R}_+$. Given the prediction $(\hat{\mu},\hat{\nu}) = f_{\bw}(x)$, we take $\hat{\mu}$ directly as the scalar prediction $\hat{y}$.

\begin{figure*}[t]
    \centering
    \includegraphics[width=\textwidth]{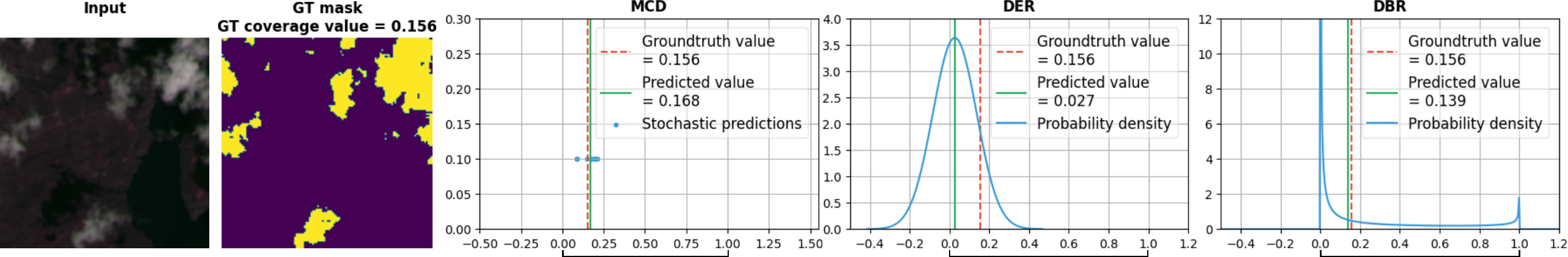}
    \caption{Qualitative comparison of MC Dropout, DER and DBR in uncertainty estimation for cloud coverage prediction. The first two panels show the input image $x$ and the foreground mask corresponding to the groundtruth coverage value $y$. The remaining panels show the three uncertainty estimation methods. Note that the predictive distribution of DER (a Gaussian) extends beyond the target range $[0,1]$, while that of DBR (a Beta) lies within $[0,1]$.}
    \label{fig:qualitative_result}
    \vspace{-15pt}
\end{figure*}

\subsection{Predictive uncertainty}

We can directly calculate the \emph{differential entropy}~\cite{michalowicz2013handbook}, which measures the predictive uncertainty of the fitted beta distribution under mean-precision parameterisation as
\begin{equation}\label{eq:dbruncertainty}
\begin{split}
  h_\bw(x) = \log\frac{\Gamma(\hat{\mu}\hat{\nu})\Gamma\bigl((1-\hat{\mu})\hat{\nu}\bigl)}{\Gamma(\hat{\nu})} \\ + (\hat{\nu}-2)\psi(\hat{\nu}) - (\hat{\mu}\hat{\nu}-1)\psi(\hat{\mu}\hat{\nu}) \\ - \bigl((1-\hat{\mu})\hat{\nu}-1\bigl)\psi\bigl((1-\hat{\mu})\hat{\nu}\bigl),
\end{split}
\end{equation}
where $\psi(\cdot)$ is the digamma function. Note that the range of~\eqref{eq:dbruncertainty} is $[-\infty,0]$. A higher entropy indicates that the predicted beta distribution has a shape closer to the uniform distribution, revealing the model is uncertain about the value of $\hat{y}$ in $[0, 1]$. Conversely, a lower entropy arises from a more concentrated density, indicating higher certainty.

In contrast to MC dropout, computing $h_\bw(x)$ in DBR depends on values $(\hat{\mu},\hat{\nu})$ obtained from a single forward pass of $f_\bw$. In contrast to DER, the predictive density respects the range $[0,1]$. Fig.~\ref{fig:qualitative_result} provides qualitative comparisons between the uncertainty estimation methods. As will be presented in Sec.~\ref{sec:results}, the groundtruth targets frequently occupy the extremes ($0$ and $1$) in real image regression datasets, thus, handling cases at the boundaries is vital.

\subsection{Architecture and training}

% Old version:
% Let $f$ be the deep neural network employed in this studey, which uses ResNet18 ~\cite{he2016deep}, MobileNetv3, or MobileNetv4 backbone in our work.

Let $f$ be the deep neural network employed in this study. Given labelled dataset $\cD  = \{ (x_j, y_j) \}^{M}_{j=1} := (\cX,\cY)$, the training of $f_\bw$ is formulated as multitask optimisation, in which both the standard regression error (root mean-squared error) and the negative log-likelihood are minimised simultaneously. Let $(\hat{\cU}, \hat{\cV}) = f_{\bw}(\cX)$, where $\hat{\cU} = \{ \hat{\mu}_j \}^{M}_{j=1}$ and $\hat{\cV} = \{ \hat{\nu}_j \}^{M}_{j=1}$. The loss is
\begin{equation}
  \cL_{s} = \cL_{RMSE}(\hat{\cU}, \cY) + \lambda\cL_{NLL}(\hat{\cU},\hat{\cV}, \cY),
  \label{eq:loss_beta_regression}
\end{equation}
where $\lambda$ weighs the relative importance of the two tasks,
\begin{align}
    \cL_{RMSE}(\hat{\cU}, \cY) = \sqrt{\sum_{j=1}^{M}\frac{(\hat{\mu}_{j}-y_{j})^2}{M}}
    \label{eq:loss_regression}
\end{align}
evaluates the regression error, while minimising
\begin{equation}
\begin{split}
  \cL_{NLL}(\hat{\cU}, \hat{\cV}, \cY) =\frac{1}{M}\sum_{j=1}^{M}\log\frac{\Gamma(\hat{\mu}_{j} \hat{\nu}_{j})\Gamma\bigl((1-\hat{\mu}_{j})\hat{\nu}_{j}\bigl)}{\Gamma(\hat{\nu}_{j})}\\ - (\hat{\mu}_{j}\hat{\nu}_{j}-1)\log(y_{j}) - \bigl((1-\hat{\mu}_{j}) \hat{\nu}_{j}-1\bigl)\log(1-y_{j})
  \label{eq:loss_NLL}
\end{split}
\end{equation}
fits the predictive beta density for uncertainty estimation. Refer to the supp material for details on the derivations.
%========================================================================
\section{Datasets}

We first describe the RS datasets used in the experiments.

% \subsubsection{Real scenes with real clouds from Landsat-8 (\texttt{RSRC-L8})} 
% \subsubsection{\texttt{RSRC-L8}: Real scenes with real clouds from Landsat-8} 

% \begin{flushleft}\textbf{}Real scenes with real clouds from Landsat-8\end{flushleft}
\subsection{Real scenes with real clouds from Landsat-8 (\texttt{RSRC-L8})} 
\begin{figure}[h]
    \centering
    \includegraphics[width=0.93\columnwidth]{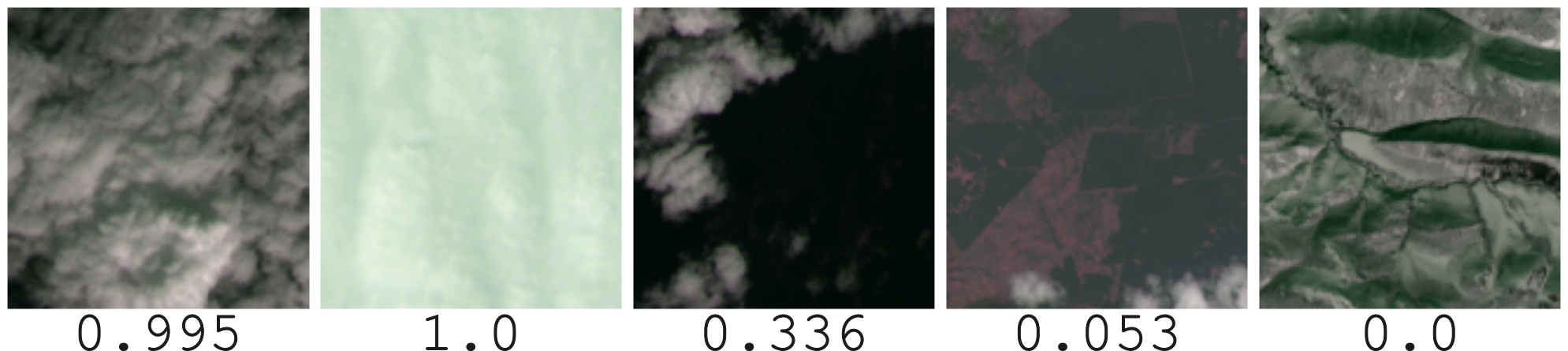}
    \caption{Sample images from \texttt{RSRC-L8} with reg.~targets.}
    \label{fig:cloud38}
    \vspace{-15pt}
\end{figure}

We used real RGB images from the 38-Cloud dataset~\cite{mohajerani2019cloud} which contained groundtruth cloud masks. We selected 18 scenes for EL and the remaining 20 scenes were used as a testing data from unseen before scenes. We cropped all scenes and their ground-truth masks into $128$x$128$ non-overlapping patches, and the cloud coverage percentage of each patch was obtained by counting the foreground pixels in the masks. In total, we have $20,294$ patches for EL, and $22,528$ for testing; see Fig.~\ref{fig:cloud38}.

% \begin{figure}[h]
%     \centering
%     \includegraphics[width=0.87\columnwidth]{imgs/5_Clouds_with_Gts_resized.png}
%     \caption{Sample images from \texttt{RSRC-L8} with reg.~targets.}
%     \label{fig:cloud38}
% \end{figure}

\subsection{Real scenes with real clouds from Sentinel-2 (\texttt{RSRC-S2})}
We sourced real base images from CloudSEN12~\cite{aybar2022cloudsen12}, a large-scale expert-labelled cloud segmentation dataset built upon ESA's Sentinel-2 L1C imagery. Due to the large size of the corpus (nearly 1 TB), we selected $5,000$ based images (about 10\%) for EL and $2,000$ based images for testing. As above, we cropped each based image into $128$x$128$ patches and used bands 4-3-2, which can be processed to RGB images. The ground-truth masks of CloudSEN12 had 4 classes: 0-clear, 1-thick cloud, 2-thin cloud, 3-cloud shadow. We grouped classes 0 \& 3 as non-cloud and classes 1 \& 2 as cloud before calculating the groundtruth cloud percentage. In total, we have $80,000$ patches for EL and $32,000$ patches for testing.

\subsection{Real scenes with land coverage (\texttt{RSLC})}
\begin{figure}[h]
    \centering
    \includegraphics[width=0.93\columnwidth]{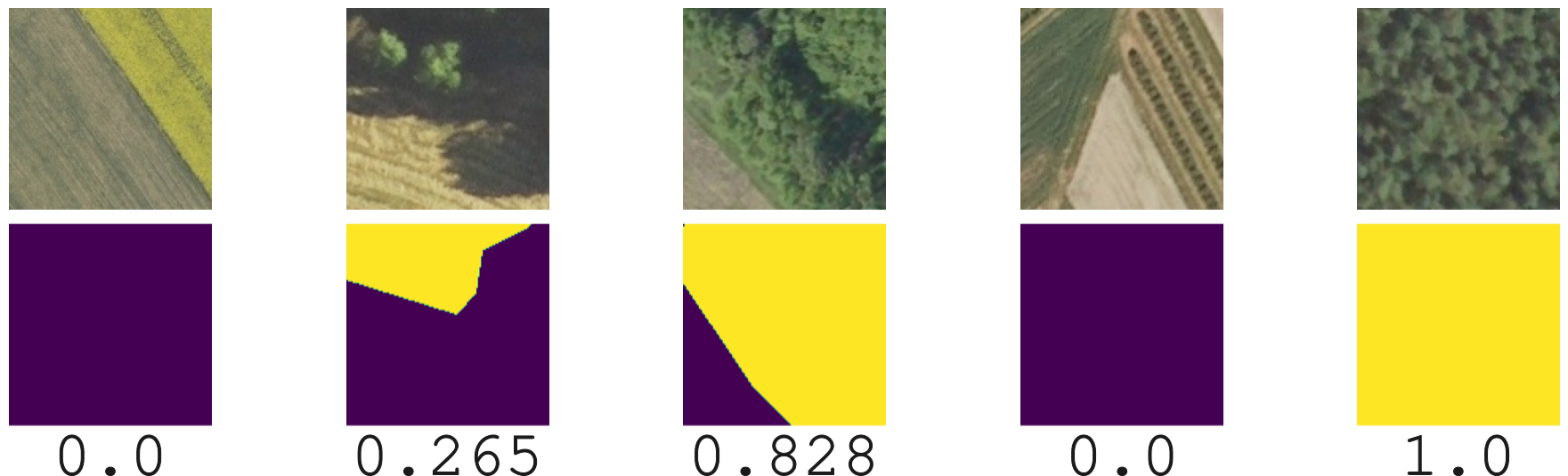}
    \caption{Sample images from \texttt{RSLC} with regression targets.}
    \label{fig:landcoverai}
\end{figure}

We sourced real RGB images for land cover segmentation from LandCover.ai~\cite{boguszewski2021landcover}. The background class (uncovered) relates to open land (\emph{e.g.}, grass, fields) while the rest (\emph{e.g.}, building, woodland, water and road) are subsumed under the foreground class (covered). Based on this assignment, the percentage of pixels corresponding to land cover were computed. As above, $128$x$128$ pixel patches were extracted, resulting in $154,056$ patches for EL and $25,279$ patches for testing; see Fig.~\ref{fig:landcoverai} for sample images.
% \begin{figure}[h]
%     \centering
%     \includegraphics[width=0.90\columnwidth]{imgs/LandCover_resized.png}
%     \caption{Sample images from \texttt{RSLC} with regression targets.}
%     \label{fig:landcoverai}
% \end{figure}
\vspace{-15pt}
\subsection{Dataset statistics} 

\begin{figure}[h]
    \centering
    \includegraphics[width=0.47\textwidth]{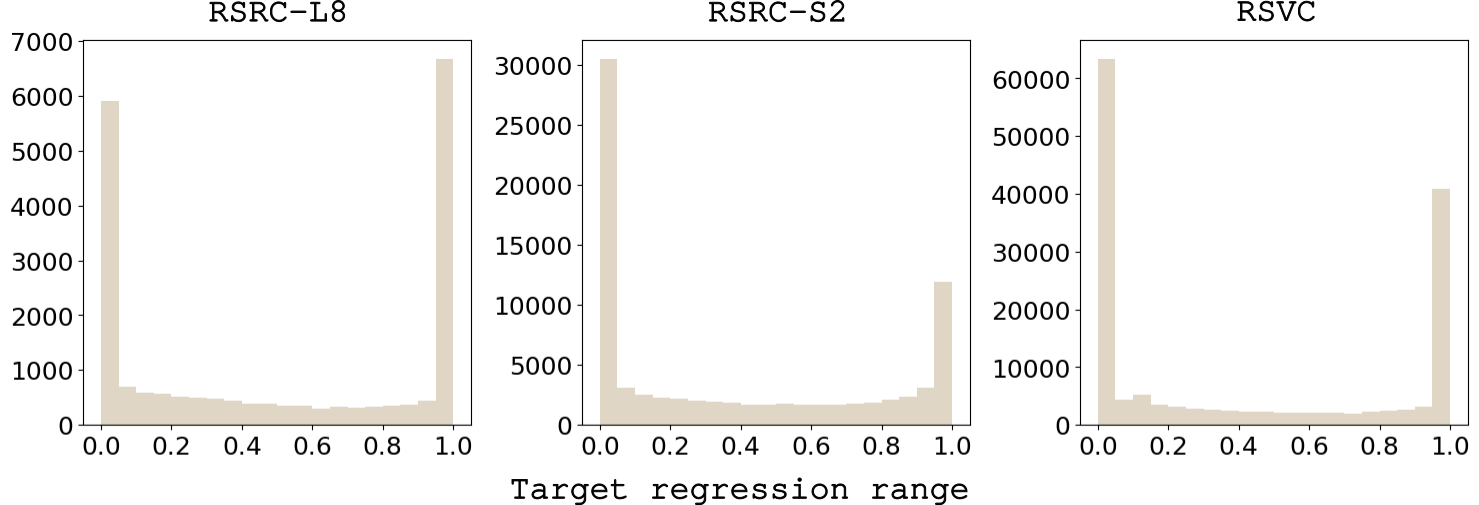}
    \caption{Histogram of groundtruth regression labels.}
    \label{fig:labeldistribution}
    \vspace{-17pt}
\end{figure}

Fig.~\ref{fig:labeldistribution} plots the histograms of the groundtruth regression labels $y$ in the three datasets above. The histograms are bimodal, with the two extremes $y=0$ and $y=1$ being much more frequent, though non-trivial number of samples span the range $[0,1]$. Crucially, the fact that many groundtruth labels occupy the extremes argues for uncertainty estimation that handles boundaries correctly.

% \begin{figure}[h]
%     \centering
%     \includegraphics[width=0.43\textwidth]{imgs/3_real_dataset_dist_font18.png}
%     \caption{Histogram of groundtruth regression labels.}
%     \label{fig:labeldistribution}
% \end{figure}

\section{Experiments}\label{sec:results} % 
\subsection{Benchmarking on RS applications}\label{subsec:benchmarking_pipelines}

We benchmarked UGEL against representative methods for model learning/updating in image-based RS applications. The specific pipelines  compared were:
\begin{itemize}
\item \citet{chen2021semi}'s SSL algorithm based on cross pseudo supervision for image semantic segmentation that we modified for image regression (predicting scalar $y$).
% \item \citet{yao2023cloud}'s ASSL algorithm that combines AL (using MC dropout) and SSL for cloud segmentation, which we also modified for image regression.
\item \citet{yao2023cloud}'s ASSL algorithm that combines AL and SSL for cloud segmentation, which we also modified for image regression.
\item The proposed UGEL with DBR uncertainty estimation. The hyperparameters used for each round of UGEL were $B_U = 6$, $B_C = |\cD|$ and $\tau = 2$.
\end{itemize}
For completeness, we also compared against
\begin{itemize}
\item Using BALD~\cite{gal2017deep} to train an image regressor. BALD is a seminal AL method that uses MC dropout for uncertainty estimation in the acquisition function that selects $\cT_U$.
\item Baseline AL method with random sampling as the acquisition function. \citet{munjal2022towards} demonstrated that random sampling performs competitively with most AL methods in fair experimental settings, validating its role as a bona fide AL baseline.
\end{itemize}
For the methods that require human labelling of $\cT_U$, we take the groundtruth labels of $\cT_U$ as the human labels.
% Old version
% We adopted the standard batch-mode active learning setup~\cite{ash2019deep}, initialising with $M=12$ labelled samples and using a batch size of $b=6$ to respect the communication limit $B_U$ for annotation. All regression models employed a ResNet18 backbone~\cite{he2016deep} and were trained with the Adam optimizer~\cite{kingma2014adam} (learning rate = 0.001). In each round, the models were randomly re-initialised and retrained for 12 epochs. Performance was evaluated using RMSE on the test set after each round. Each experiment consisted of as many rounds as feasible on our HPC, and the RMSE values were recorded. To ensure robustness and comparability, experiments were repeated 10 times to obtain mean, std dev and p-values. Results on \texttt{RSRC-L8} and \texttt{RSLC} are in Fig.~\ref{fig:cloudL8_ablation_v3}, while results on \texttt{RSRC-S2} are in the supp material.

\begin{figure*}[ht]\centering
% ResNet18
\includegraphics[height=0.170\textwidth]{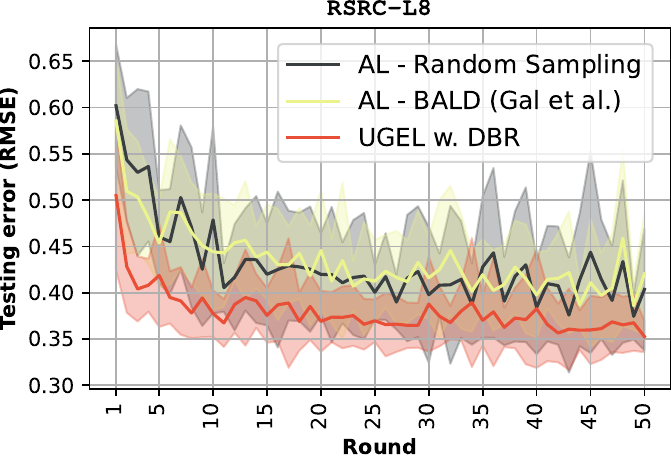}
\includegraphics[height=0.169\textwidth]{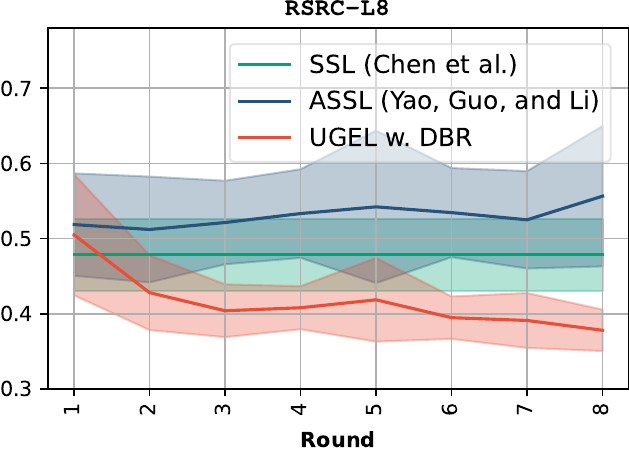}
\includegraphics[height=0.170\textwidth]{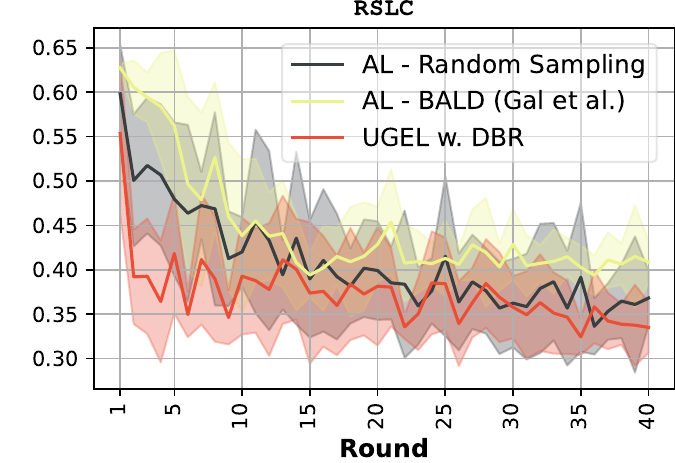}
\includegraphics[height=0.169\textwidth]{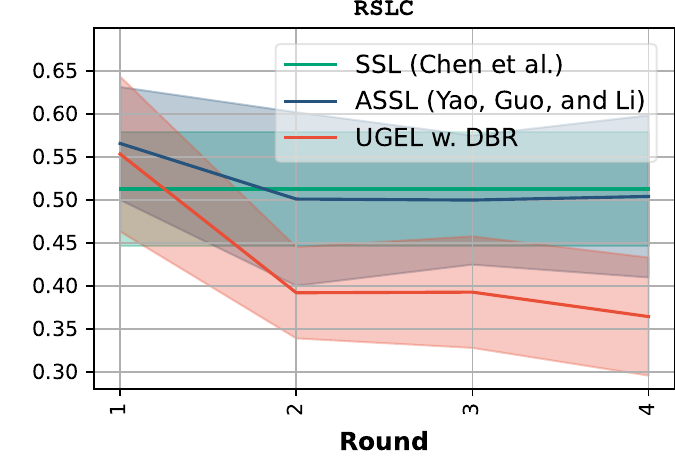}
% MobileNetv3
\includegraphics[height=0.170\textwidth]{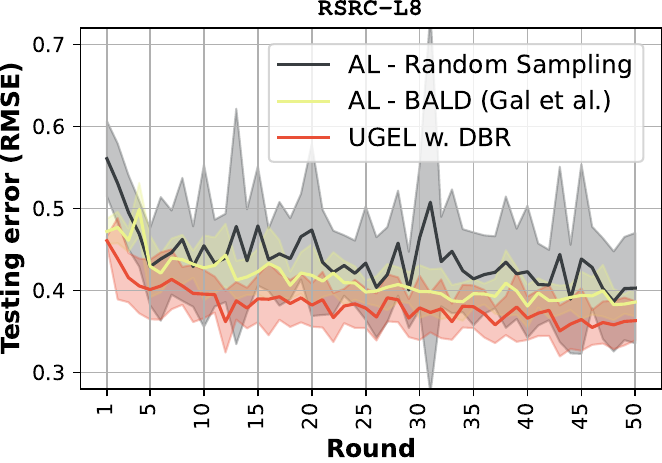}
\includegraphics[height=0.169\textwidth]{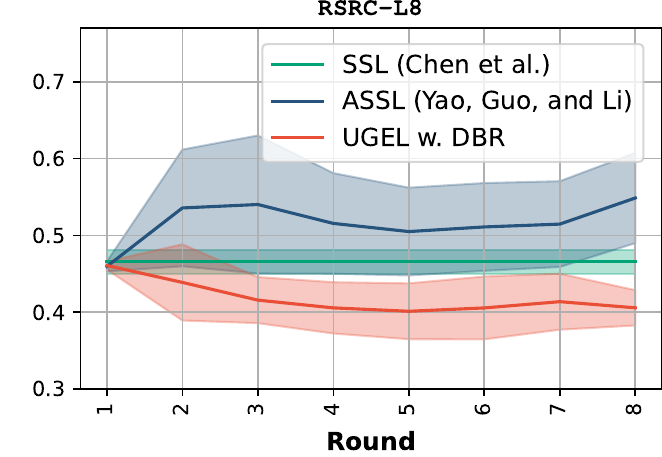}
\includegraphics[height=0.170\textwidth]{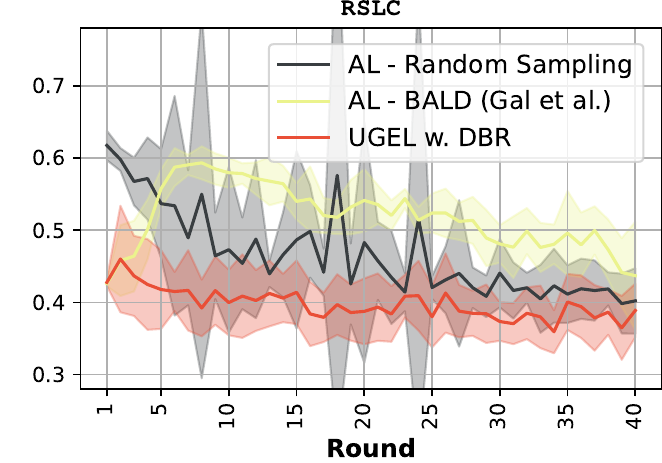}
\includegraphics[height=0.169\textwidth]{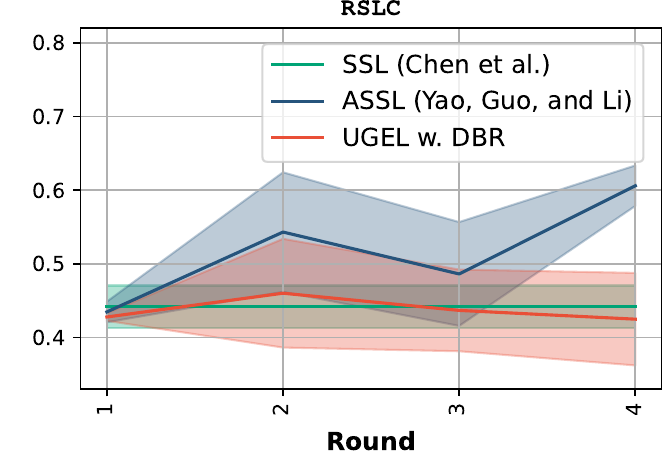}
\caption{Panels 1--4 (top row): Benchmarking UGEL with DBR against AL (BALD~\cite{gal2017deep} and random sampling), SSL~\cite{chen2021semi} and ASSL~\cite{yao2023cloud} on \texttt{RSRC-L8} and \texttt{RSLC} using a ResNet18 backbone.  Panels 5--8 (bottom row): Same comparisons but using MobileNetV3 as the backbone. Bolded curves and shaded regions respectively indicate mean RMSE testing error and standard deviation over 10 runs.}    
    \label{fig:benchmarking}
    \vspace{-9pt}
\end{figure*}

We adopted the standard batch-mode active learning setup~\cite{ash2019deep}, initialising with $M=12$ labelled samples and using a batch size of $b=6$ to respect the communication limit $B_U$ for annotation. $B_U$ is kept small to reflect the practical constraints of satellite communication. As discussed in Sec.~\ref{sec:intro}, satellites operate under stringent computational and energy limitations, which necessitates the use of lightweight deep neural networks. Accordingly, this work focuses on compact yet effective architectures. All regression models employed widely adopted lightweight backbone architectures (ResNet18~\cite{he2016deep}, MobileNetV3~\cite{howard2019searching}, or MobileNetV4~\cite{qin2024mobilenetv4}) and were trained with the Adam optimiser~\cite{kingma2014adam} (learning rate = 0.001). In each round, the models were randomly re-initialised and retrained for 12 epochs. Re-initialisation at the beginning of each round is required, since continuing training from the previous checkpoint degrades model performance due to catastrophic forgetting. Performance was evaluated using RMSE on the test set after each round. Each experiment consisted of as many rounds as feasible on our HPC, and the RMSE values were recorded. To ensure robustness and comparability, experiments were repeated 10 times to obtain mean, std dev and p-values. Results on \texttt{RSRC-L8} and \texttt{RSLC} are presented in Fig.~\ref{fig:benchmarking} for models using ResNet18 and MobileNetV3. Results on \texttt{RSRC-S2} and for the MobileNetV4 backbone are included in the supp material. The SSL and ASSL experiments were limited to fewer rounds due to computational constraints. Each ASSL round processes all remaining unlabeled samples, greatly increasing computational cost and runtime compared with UGEL with DBR or active learning methods (Rand and BALD). Extending ASSL further would require substantially more training time and resources without providing additional methodological insight.
% Results on \texttt{RSRC-L8} and \texttt{RSLC} are in Fig.~\ref{fig:cloudL8_ablation_v3}, while results on \texttt{RSRC-S2} are in the supp material.

AL with random sampling and BALD reduced the testing RMSE over time,  but with large fluctuations, particularly for random sampling. Moreover, BALD did not outperform random sampling; in fact, in \texttt{RSLC} random sampling reduced the RMSE faster than BALD. This observation is consistent with~\cite{munjal2022towards}. In contrast to the AL methods, UGEL with DBR consistently yielded lower RMSE and standard deviations, indicating that incorporating supervisory signals from the most certain unlabelled instances enhances model convergence and learning stability.

UGEL with DBR substantially outperformed \citet{yao2023cloud}'s ASSL method. While ASSL initially provided noticeable improvement over using AL alone, its performance plateaued thereafter, likely due to a large number of noisy pseudo-labels diluting the human-labelled samples. In contrast, UGEL sustained the benefit of SSL by leveraging only the most certain pseudo-labels, resulting in consistent performance gains. This highlights the importance of selectively integrating pseudo-labels guided by uncertainty. 

% Old version
% Note that the gap between all methods reduce with the number of rounds, since asymptotically all the new data are eventually selected for labelling. The key differentiating factor for EL is reducing the testing error using the least number of samples to respect labelling and bandwidth limits. Tab.~\ref{tab:pvaluesbenchmark} shows p-values from comparing UGEL with the other methods. The performance gaps are meaningful since most of the p-values are $< 0.05$. Note that outcomes in different number of rounds are shown since the datasets are of distinct sizes, which yielded different convergence speeds.
Note that the gap between all methods reduce with the number of rounds, since asymptotically all the new data are eventually selected for labelling. The key differentiating factor for EL is reducing the testing error using the least number of samples to respect labelling and bandwidth limits. Tab.~\ref{tab:pvaluesbenchmark} shows p-values from comparing UGEL with the other methods, complementing the results shown in Fig.~\ref{fig:benchmarking}. The performance gaps are meaningful since most of the p-values are $< 0.05$. Note that outcomes in different number of rounds are shown since the datasets are of distinct sizes, which yielded different convergence speeds.

% 02/08: Merged table
\begin{table}[ht]
   \small
   \centering
   \begin{tabular}{l||lccccr}
   \toprule\toprule
   \textbf{} & \textbf{Test} & \textbf{Rd. 2} & \textbf{Rd. 4} & \textbf{Rd. 6} & \textbf{Rd. 8} \\ 
   \midrule
        \multirow{4}{*}{\rotatebox[origin=c]{90}{\scriptsize\texttt{RSRC-S2}}} & UGEL vs Rand & 0.019 & 0.001 & 0.005 & 0.002 \\
                                                                                & UGEL vs BALD & 0.042 & 0.002 & 0.053 & 0.002 \\
                                                                                & UGEL vs SSL & 0.014 & 0.001 & 0.005 & 0.001 \\
                                                                                & UGEL vs ASSL & 0.016 & 0.008 & 0.008 & 0.008 \\
    \toprule\
     \textbf{} & \textbf{Test} & \textbf{Rd. 5} & \textbf{Rd. 10} & \textbf{Rd. 15} & \textbf{Rd. 20} \\ 
        \midrule
        \multirow{4}{*}{\rotatebox[origin=c]{90}{\scriptsize\texttt{RSRC-L8}}} & UGEL vs Rand & 0.024 & 0.001 & 0.002 & 0.009 \\
                                                                                & UGEL vs BALD & 0.036 & 0.002 & 0.002 & 0.001 \\
                                                                                & UGEL vs SSL & 0.001 & 0.001 & 0.001 & 0.001\\
                                                                                & UGEL vs ASSL & 0.001 & 0.001 & 0.001 & 0.001 \\
    \toprule\
     \textbf{} & \textbf{Test} & \textbf{Rd. 1} & \textbf{Rd. 2} & \textbf{Rd. 3} & \textbf{Rd. 4} \\ 
        \midrule                                                           
        \multirow{4}{*}{\rotatebox[origin=c]{90}{\scriptsize\texttt{RSLC}}} & UGEL vs Rand & 0.188 & 0.001 & 0.005 & 0.003 \\
                                                                                & UGEL vs BALD & 0.002 & 0.001 & 0.001 & 0.001 \\
                                                                                & UGEL vs SSL &  0.839 & 0.001 & 0.001 & 0.001 \\
                                                                                & UGEL vs ASSL & 0.577 & 0.042 & 0.024 & 0.019 \\
   \bottomrule
   \end{tabular}
   \caption{Wilcoxon signed-rank test p-values comparing UGEL with DBR against other methods on all datasets using the ResNet18 backbone.}
   \label{tab:pvaluesbenchmark}
   \vspace{-15pt}
\end{table}

% =====

%---------------------------------
\subsection{UGEL ablation studies}

Two ablation tests were conducted:
\begin{itemize}
    \item[\textbf{T1}] Effects of diff.~uncertainty estimation methods on UGEL.
    \item[\textbf{T2}] Effects of different groundtruth regression label distributions on the performance of UGEL.
\end{itemize}

\begin{figure}[h]
    \centering
    \includegraphics[width=0.36\textwidth]{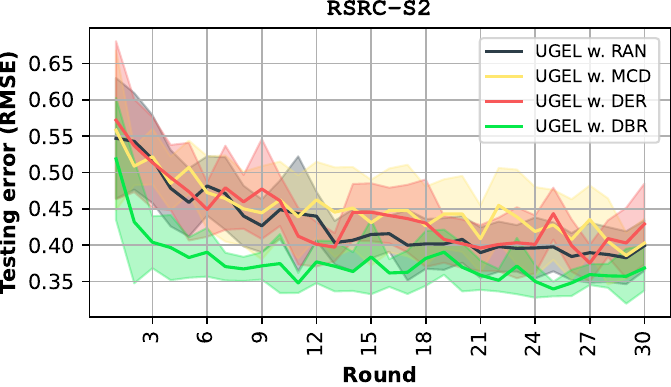}
    \caption{Comparison of different uncertainty estimation methods in UGEL on \texttt{RSRC-S2} using the ResNet18 backbone. Bolded curves and shaded regions respectively indicate mean RMSE testing error and std dev over 10 runs.}
    \label{fig:ugel_ablation}
    \vspace{-15pt}
\end{figure}

For \textbf{T1}, in place of DBR, we used the following uncertainty estimation algorithms in UGEL: MC dropout (MCD) with $P = 10$, DER and a ``dummy'' method (RAN) that assigns constant uncertainty thus amounting to randomly sampling the data. The experimental setup, hyperparameter configurations, and evaluation metric were kept consistent with those described in the preceding benchmarking experiment. The utilised backbone is ResNet18. Results on \texttt{RSRC-S2} are in Fig.~\ref{fig:ugel_ablation}, while those for \texttt{RSRC-L8} and \texttt{RSLC} are in the supp material.

The results show that using DBR in UGEL outperformed the other uncertainty methods, with a noticeable gap in the earlier rounds. The statistical significance test results in Tab.~\ref{tab:pvaluesuncertainty} confirm that using DBR in UGEL achieved significantly lower regression errors than other methods most of the time (p-values $< 0.05$). %This underscores the effectiveness of modeling predictive uncertainty using beta distribution in the RS image regression tasks.

% The results show that using DBR in UGEL outperformed the other uncertainty methods, with a noticeable gap in the earlier rounds. The statistical significance test results confirm that using DBR in UGEL achieved significantly lower regression errors than other methods most of the time (p-values $< 0.05$). This underscores the effectiveness of modeling predictive uncertainty using beta distribution in the RS image regression tasks. Table of these tests are included in the supp material. 

\begin{table}[h]
   \small
   \centering
   \begin{tabular}{l||cccccr}
   \toprule\toprule
   \textbf{} & \textbf{Test} & \textbf{Rd. 5} & \textbf{Rd. 10} & \textbf{Rd. 15} & \textbf{Rd. 20} \\ 
   \midrule
        \multirow{3}{*}{\rotatebox[origin=c]{90}{\scriptsize\texttt{RSRC-S2}}} & DBR vs RAN & 0.005 & 0.002 & 0.116 & 0.019 \\
                                                                    & DBR vs MCD & 0.001 & 0.003 & 0.08 & 0.003 \\
                                                                    & DBR vs DER  & 0.001 & 0.001 & 0.005 & 0.032 \\ 
        \midrule
        \multirow{3}{*}{\rotatebox[origin=c]{90}{\scriptsize\texttt{RSRC-L8}}} & DBR vs RAN & 0.094 & 0.003 & 0.003 & 0.068 \\
                                                                    & DBR vs MCD & 0.003 & 0.001 & 0.004 & 0.003\\
                                                                    & DBR vs DER & 0.244 & 0.003 & 0.011 & 0.002\\
        \midrule                                                           
        \multirow{3}{*}{\rotatebox[origin=c]{90}{\scriptsize\texttt{RSLC}}} & DBR vs RAN  & 0.348 & 0.423 & 0.216 & 0.348 \\
                                                                    & DBR vs MCD & 0.042 & 0.005 & 0.116 & 0.042\\
                                                                    & DBR vs DER  & 0.138 & 0.042 & 0.065 &  0.042\\
                 
   \bottomrule
   \end{tabular}
   \caption{Wilcoxon signed-rank test p-values comparing using DBR and other methods using the ResNet18 backbone, sampled every 5 rounds.}
   \label{tab:pvaluesuncertainty}
   \vspace{-15pt}
\end{table}

For \textbf{T2}, we generated and added synthetic cloud masks on clearsky patches on RS images to produce deep regression datasets with different groundtruth label distributions. UGEL with the different uncertainty estimation methods was then executed on the datasets, and the results show that DBR still provided the best regression accuracy with high statistical significance. Due to page limits, we refer the reader to the supp material for details of this experiment.

% ------------------------------
\subsection{Runtime comparisons}
To assess runtime in a satellite-borne EL context, we executed UGEL with different uncertainty methods on an Nvidia Jetson Orin NX, which was identified as viable for satellite operation \cite{Felix2024total,Rodriguez-Ferrandez2024proton}. The employed  backbone is ResNet18. Tab.~\ref{tab:example_multirow} records the runtime on \texttt{RSRC-S2} (see the supp material for the runtimes on \texttt{RSRC-L8} and \texttt{RSLC}), where it is clear that the computational cost of uncertainty estimation was much greater than that of model retraining, and that MC dropout was costlier than DER and DBR.

\begin{table}[H]
   \small
   \centering
   \begin{tabular}{lccr}
   \toprule\toprule
   \textbf{Method} & \textbf{Uncertain.~est.}$\downarrow$ & \textbf{Model retrain.}$\downarrow$  \\ 
   \midrule
            UGEL w.~RAN & n/a & 7.05-11.64\\
            UGEL w.~MCD & 495.84-514.18 & 7.18-12.54\\
            UGEL w.~DER & 82.01-85.47 & 7.23-13.29\\
            UGEL w.~DBR & 68.04-74.53 & 7.43-12.02\\
            
   \bottomrule
   \end{tabular}
   \caption{Runtime (minimum and maximum, in seconds) for different uncertainty estimation methods during the first 10 rounds of UGEL on the \texttt{RSRC-S2} dataset using the ResNet18 backbone.}
   \label{tab:example_multirow}
   \vspace{-9pt}
\end{table}

\section{Conclusions and limitations}
We proposed an uncertainty-guided edge learning (UGEL) algorithm that was integrated with an efficient uncertainty estimation method based on deep beta regression (DBR). Benchmarking on real remote sensing (RS) datasets demonstrated the superiority of UGEL with DBR over active and semi-supervised learning methods. Ablation studies confirmed that DBR effectively guided the selection of informative unlabeled samples for model updates in remote sensing applications. Overall, the results indicate that the proposed framework is well-suited for accelerating on-board deep network training for RS applications.

% The gains from the proposed framework decrease as object diversity increases, as evidenced by the \texttt{RSLC} results. In addition, the twin-model design in UGEL introduces extra computational, memory, and power overhead during training. Future work will explore more adaptive uncertainty estimation for diverse settings and more efficient semi-supervised methods that avoid using a twin model.

The improvement provided by the proposed framework declines with increased object diversity, as shown in the results on \texttt{RSLC}. The use of the twin model in UGEL also increases computational costs, memory usage, and power consumption during training. In the future, we plan to develop a more adaptive uncertainty estimation method for such conditions and a more efficient semi-supervised learning method without employing a twin model.

\vfill

%Acknowledgements
\vspace{-5pt}
\section*{Acknowledgments}
% \vspace{-5pt}
% We thank Andrew Du, Anh-Dzung Doan, and Punarjay Chakravarty for helpful discussions and acknowledge the computational resources provided by the Phoenix HPC at Adelaide University.
We thank Andrew Du, Anh-Dzung Doan, and Punarjay Chakravarty for insightful discussions and feedback. %We gratefully acknowledge the computational resources provided by the Phoenix HPC at the Adelaide University.

{
    \small
    \bibliographystyle{ieeenat_fullname}
    \bibliography{main}
}

\clearpage
\setcounter{page}{1}
\maketitlesupplementary

\section{Derivations}\label{sec:derivation}
We have the likelihood of a beta distribution given the mean $\mu$ and the precision $\nu$:
\begin{equation}
  p(y|\mu, \nu) = \frac{\Gamma(\nu)}{\Gamma(\mu\nu)\Gamma((1-\mu)\nu)}y^{\mu\nu-1}(1-y)^{(1-\mu)\nu-1},
  \label{eq:likelihood}
\end{equation}
where $\Gamma(\cdot)$ is the gamma function. We can compute the negative log likelihood loss, $\cL_{NLL}$, as:
\begin{equation}
\begin{split}
\cL_{NLL}(\mu, \nu, y)  &= -\log{p(y|\mu, \nu)}\\
&= -[\log{\frac{\Gamma(\nu)}{\Gamma(\mu\nu)\Gamma((1-\mu)\nu)}} + \log{y^{\mu\nu-1}}
\\&\quad+ \log{(1-y)^{(1-\mu)\nu-1}}]
\\&= \log{\frac{\Gamma(\mu\nu)\Gamma((1-\mu)\nu)}{\Gamma(\nu)}}  - (\mu\nu-1)\log{y}
\\&\quad+ ((1-\mu)\nu-1)\log{(1-y)}
\end{split}
\end{equation}\label{eq:nll}
The differential entropy of the beta distribution, given shape parameters $\alpha,\beta >0$, is commly expressed as:
\begin{equation}
\begin{split}
h &= \log{Beta(\alpha, \beta)} - (\alpha-1)[\psi(\alpha)-\psi(\alpha+\beta)]
\\&\quad- (\beta-1)[\psi(\beta)-\psi(\alpha+\beta)]
\end{split}
\end{equation}\label{eq:d_entropy}
where $\psi(\cdot)$ is the digamma function. In this work, we use an alternative parameterisation with the mean $\mu = \alpha/(\alpha + \beta)\in(0,1)$ and the precision $\nu = \alpha + \beta>0$. Therefore, the parameters $\alpha$ and $\beta$ can be represented in terms of the mean $\mu$ and precision $\nu$:
\begin{equation}
\begin{split}
\alpha &= \mu\nu,\\
\beta &= (1-\mu)\nu
\end{split}
\end{equation}\label{eq:d_entropy}
Thereby, the differential entropy of the beta distribution is calculated as:
\begin{equation}
\begin{split}
h &= \log{Beta(\mu\nu, (1-\mu)\nu)}
\\&\quad - (\mu\nu-1)[\psi(\mu\nu)-\psi(\mu\nu+(1-\mu)\nu))]
\\&\quad- ((1-\mu)\nu-1)[\psi((1-\mu)\nu)-\psi(\mu\nu+(1-\mu)\nu)]
\\&=\log\frac{\Gamma(\mu\nu)\Gamma((1-\mu)\nu)}{\Gamma(\nu)}
\\&\quad+ (\nu-2)\psi(\nu) - (\mu\nu-1)\psi(\mu\nu)
\\&\quad- ((1-\mu)\nu-1)\psi((1-\mu)\nu)
\\&= \log{\Gamma(\mu\nu)}+\log(\Gamma((1-\mu)\nu))-\log{\Gamma(\nu)}
\\&\quad- (\mu\nu-1)[\psi(\mu\nu)-\psi(\nu)]
\\&\quad-((1-\mu)\nu-1)[\psi((1-\mu)\nu) - \psi(\nu)]
\\&= \log{\Gamma(\mu\nu)}+\log(\Gamma((1-\mu)\nu))-\log{\Gamma(\nu)}
\\&\quad+(\nu-2)\psi(\nu) - (\mu\nu-1)\psi(\mu\nu)
\\&\quad- ((1-\mu)\nu-1)\psi((1-\mu)\nu)
\end{split}
\end{equation}\label{eq:d_entropy}

\section{Benchmarking on RS applications}
\begin{figure}[ht]\centering
\includegraphics[height=0.493\columnwidth]{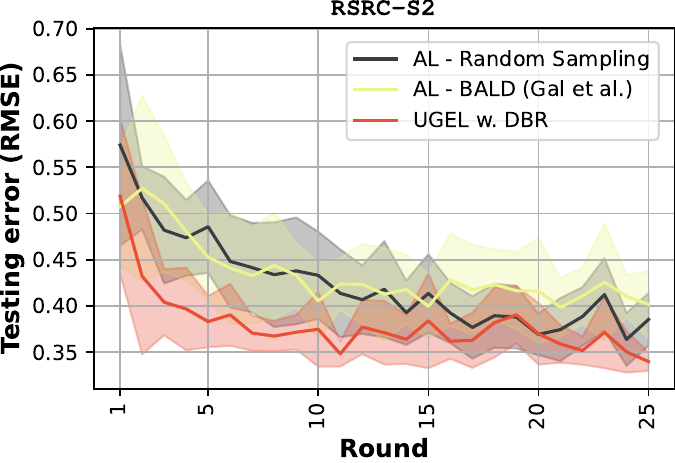}
\includegraphics[height=0.493\columnwidth]{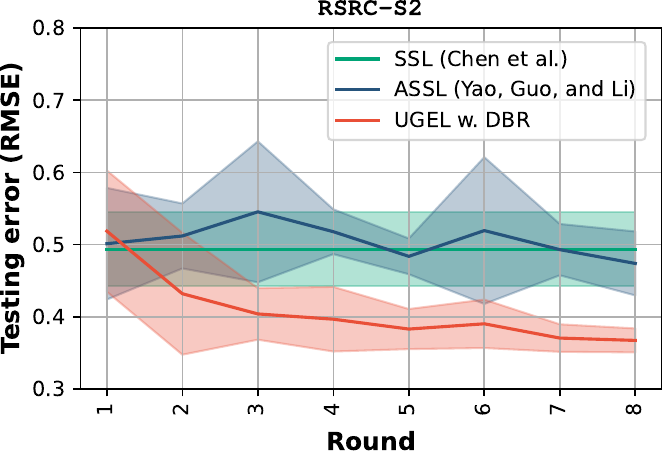}
\caption{Benchmarking UGEL with DBR against AL (BALD and random sampling), SSL and ASSL on \texttt{RSRC-S2} using the ResNet18 backbone.  Bolded curves and shaded regions respectively indicate mean RMSE testing error and std dev over 10 runs.\\}    
    \label{fig:cloudSEN12_ablation_resnet18}
\end{figure}

\begin{figure}[ht]\centering
\includegraphics[height=0.493\columnwidth]{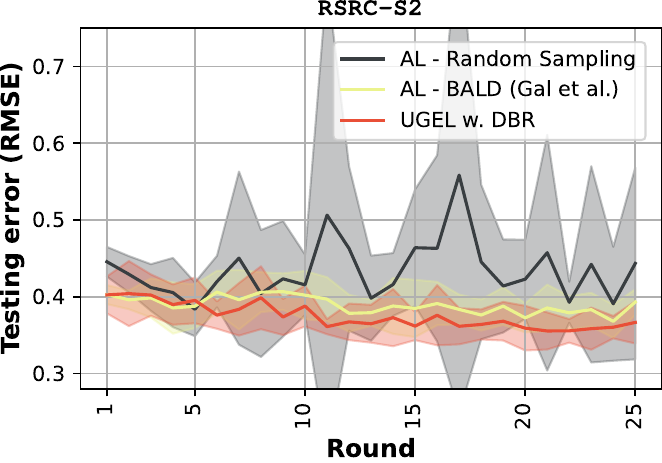}
\includegraphics[height=0.493\columnwidth]{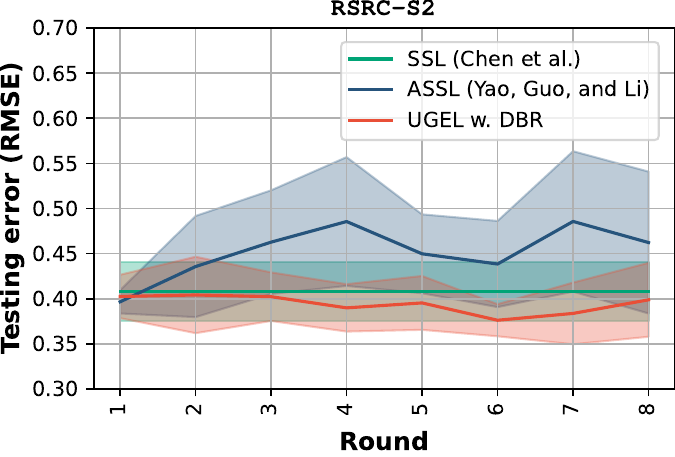}
\caption{Benchmarking UGEL with DBR against AL (BALD and random sampling), SSL and ASSL on \texttt{RSRC-S2} using the MobileNetV3 backbone.  Bolded curves and shaded regions respectively indicate mean RMSE testing error and std dev over 10 runs.}    
    \label{fig:cloudSEN12_ablation_mobilenetv3}
\end{figure}

\begin{figure}[ht]\centering
\includegraphics[height=0.495\columnwidth]{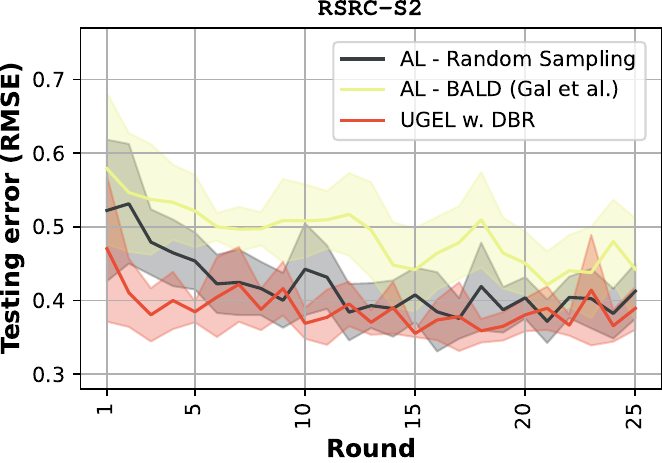}
\includegraphics[height=0.495\columnwidth]{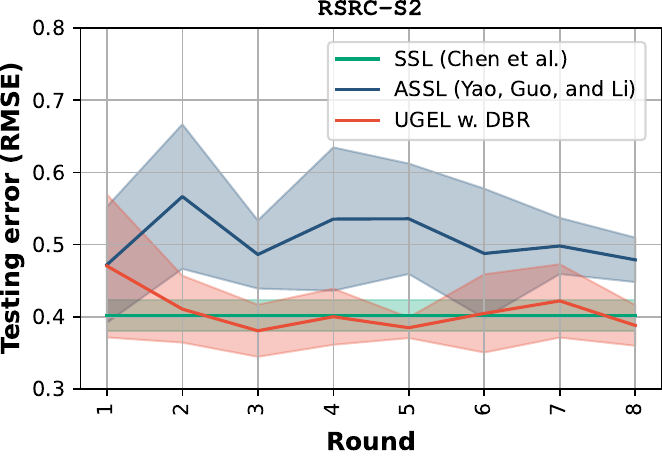}
\caption{Benchmarking UGEL with DBR against AL (BALD and random sampling), SSL and ASSL on \texttt{RSRC-S2} using the MobileNetV4 backbone.  Bolded curves and shaded regions respectively indicate mean RMSE testing error and std dev over 10 runs.}    
    \label{fig:cloudSEN12_ablation_mobilenetv4}
\end{figure}

\begin{figure}[ht]\centering
\includegraphics[height=0.493\columnwidth]{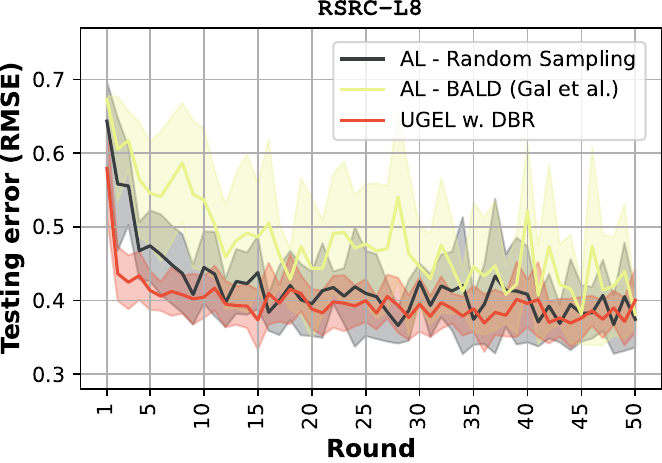}
\includegraphics[height=0.493\columnwidth]{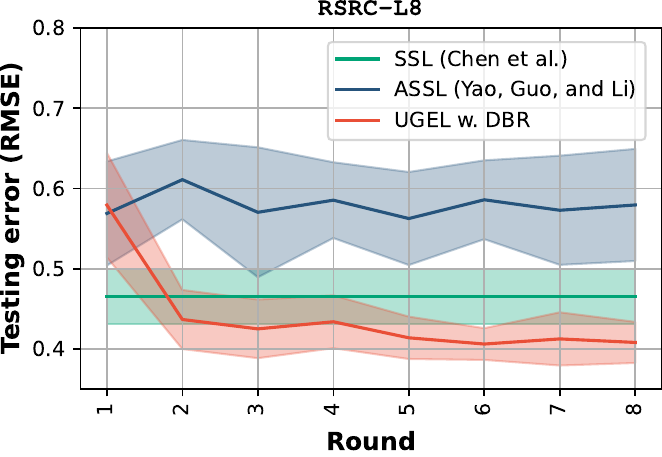}
\caption{Benchmarking UGEL with DBR against AL (BALD and random sampling), SSL and ASSL on \texttt{RSRC-L8} using the MobileNetV4 backbone.  Bolded curves and shaded regions respectively indicate mean RMSE testing error and std dev over 10 runs.}    
    \label{fig:cloudL8_benchmark_mobilenetv4}
\end{figure}

\begin{figure}[ht]\centering
\includegraphics[height=0.495\columnwidth]{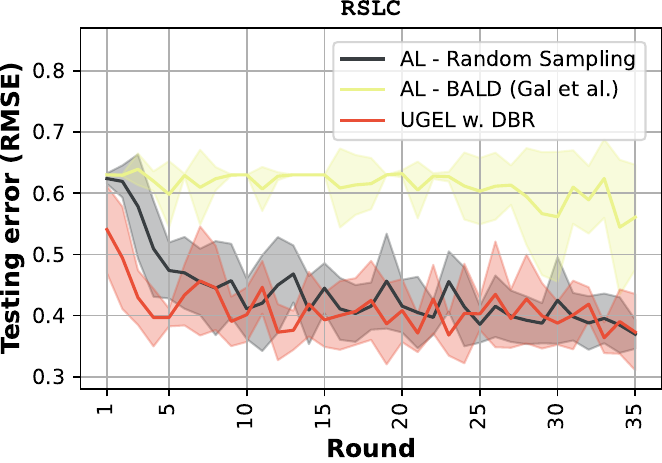}
\includegraphics[height=0.495\columnwidth]{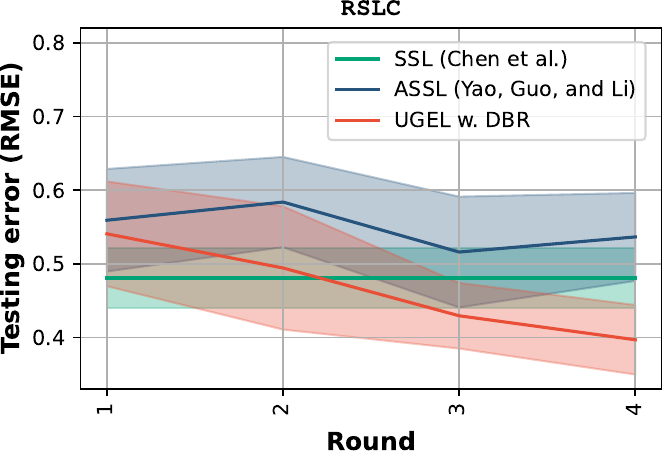}
\caption{Benchmarking UGEL with DBR against AL (BALD and random sampling), SSL and ASSL on \texttt{RSLC} using the MobileNetV4 backbone.  Bolded curves and shaded regions respectively indicate mean RMSE testing error and std dev over 10 runs.}    
    \label{fig:land_benchmark_mobilenetv4}
\end{figure}

Fig.~\ref{fig:cloudSEN12_ablation_resnet18}, Fig.~\ref{fig:cloudSEN12_ablation_mobilenetv3}, and Fig.~\ref{fig:cloudSEN12_ablation_mobilenetv4} present the benchmarking results of UGEL against representative methods on \texttt{RSRC-S2} using the ResNet18, MobileNetV3, and MobileNetV4 backbones, respectively. These results are consistent with those observed on \texttt{RSRC-L8} and \texttt{RSLC}, as shown in Fig.~\ref{fig:benchmarking} in the main paper. Additionally, Fig.~\ref{fig:cloudL8_benchmark_mobilenetv4} and Fig.~\ref{fig:land_benchmark_mobilenetv4} present the benchmarking results of UGEL against representative methods on \texttt{RSRC-L8} and \texttt{RSLC} using the MobileNetV4 backbone.

Tab.~\ref{tab:pvaluesbenchmark_mobilenetv3} and Tab.~\ref{tab:pvaluesbenchmark_mobilenetv4} provide the Wilcoxon signed-rank test p-values comparing UGEL with DBR to other methods using the MobileNetV3 and MobileNetV4 backbones, respectively.

% ResNet18

% MobileNetv3

% MobileNetv4
% RSRC-S2

% RSRC-L8

% RSLC
% =============================
\section{UGEL ablation studies}
\subsection{[\textbf{T1}] Effects of different uncertainty estimation methods on UGEL.}

\begin{figure}[ht]
    \centering
    \includegraphics[width=0.36\textwidth]{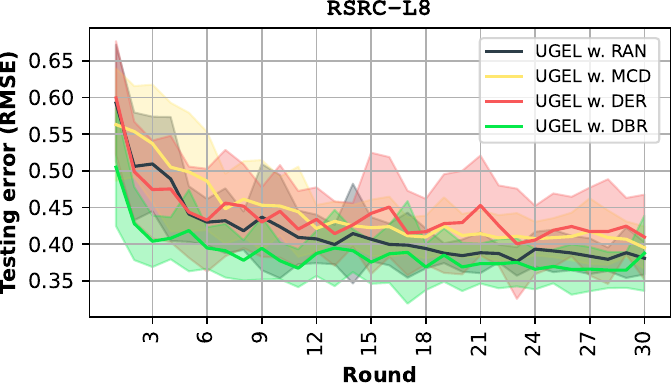}
    \includegraphics[width=0.36\textwidth]{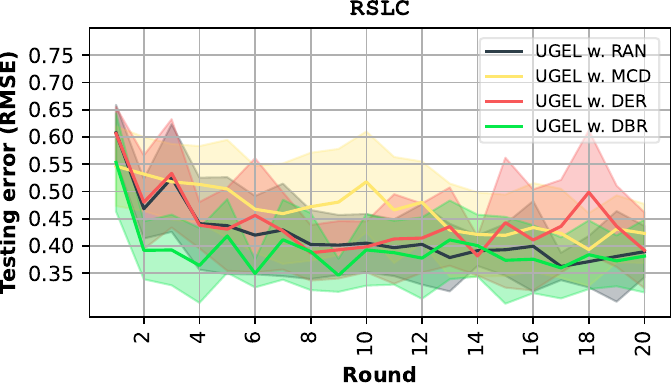}
    \caption{(Top) Comparison of different uncertainty estimation methods in UGEL on \texttt{RSRC-L8} using the ResNet18 backbone. (Bottom) Same comparison on \texttt{RSLC} using the ResNet18 backbone. Bolded curves and shaded regions respectively indicate mean RMSE testing error and std dev over 10 runs.}
    \label{fig:ugel_ablation_rsrcl8_rslc}
\end{figure}

The experimental setup, hyperparameter configurations, and evaluation metric were kept consistent with those described in the main paper. We evaluated the impact of different uncertainty estimation methods within the UGEL framework. In addition to Fig.~\ref{fig:ugel_ablation} in the main paper, Fig.~\ref{fig:ugel_ablation_rsrcl8_rslc} provides an empirical comparison of their strengths and limitations in cloud and land coverage prediction tasks.

The results demonstrate that DBR is significantly more effective than other uncertainty methods in UGEL, particularly in the early rounds where the performance gap was most evident. DBR's uncertainty estimations facilitated the selection of informative unlabeled samples for model updating. On \texttt{RSRC-L8} and \texttt{RSRC-S2} datasets, DBR achieved not only lowest RMSE but also the smallest standard deviation, indicating greater reliability for real-world cloud coverage prediction. For the land coverage prediction using \texttt{RSLC}, while DBR still outperformed in early rounds, DER and RAN quickly reached comparable performance. This outcome may be attributed to the greater diversity of objects considered as coverage in the land coverage prediction task.

% MobileNetv3
\begin{table}[ht]
   \small
   \centering
   \begin{tabular}{l||lccccr}
   \toprule\toprule
   \textbf{} & \textbf{Test} & \textbf{Rd. 2} & \textbf{Rd. 4} & \textbf{Rd. 6} & \textbf{Rd. 8} \\ 
   \midrule
        \multirow{4}{*}{\rotatebox[origin=c]{90}{\scriptsize\texttt{RSRC-S2}}} & UGEL vs Rand & 0.080 & 0.348 & 0.014 & 0.688 \\
                                                                                & UGEL vs BALD & 0.688 & 0.722 & 0.032 & 0.312 \\
                                                                                & UGEL vs SSL & 0.461 & 0.065 & 0.042 & 0.348 \\
                                                                                & UGEL vs ASSL & 0.116 & 0.002 & 0.007 & 0.019 \\
    \toprule\
     \textbf{} & \textbf{Test} & \textbf{Rd. 2} & \textbf{Rd. 4} & \textbf{Rd. 6} & \textbf{Rd. 8} \\ 
        \midrule
        \multirow{4}{*}{\rotatebox[origin=c]{90}{\scriptsize\texttt{RSRC-L8}}} & UGEL vs Rand & 0.007 & 0.001 & 0.065 & 0.053 \\
                                                                                & UGEL vs BALD & 0.01 & 0.001 & 0.246 & 0.024 \\
                                                                                & UGEL vs SSL & 0.053 & 0.005 & 0.005 & 0.001 \\
                                                                                & UGEL vs ASSL & 0.01 & 0.002 & 0.001 & 0.001 \\
    \toprule\
     \textbf{} & \textbf{Test} & \textbf{Rd. 1} & \textbf{Rd. 2} & \textbf{Rd. 3} & \textbf{Rd. 4} \\ 
        \midrule                                                           
        \multirow{4}{*}{\rotatebox[origin=c]{90}{\scriptsize\texttt{RSLC}}} & UGEL vs Rand & 0.001 & 0.001 & 0.001 & 0.001 \\
                                                                                & UGEL vs BALD & 0.754 & 0.577 & 0.161 & 0.014 \\
                                                                                & UGEL vs SSL &  0.161 & 0.754 & 0.423 & 0.313\\
                                                                                & UGEL vs ASSL & 0.406 & 0.156 & 0.062 & 0.031 \\
   \bottomrule
   \end{tabular}
   \caption{Wilcoxon signed-rank test p-values comparing UGEL with DBR against other methods on all datasets using the MobileNetV3 backbone.}
   \label{tab:pvaluesbenchmark_mobilenetv3}
\end{table}

\subsection{[\textbf{T2}] Effects of different groundtruth regression label distributions on the performance of UGEL.}

\begin{figure}[ht]
    \centering
    \includegraphics[width=0.99\columnwidth]{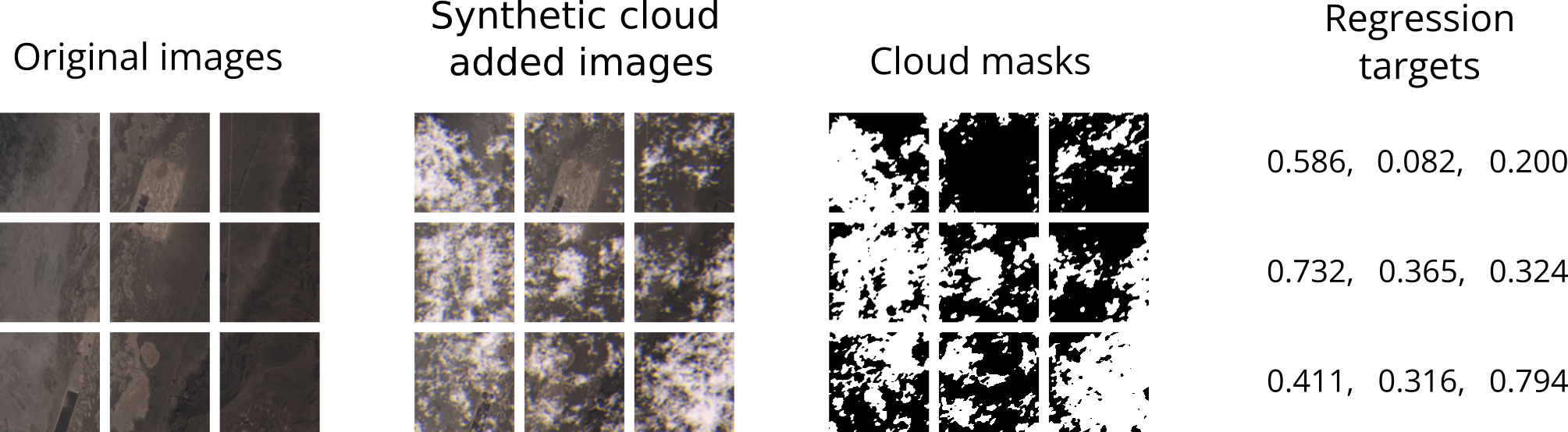}
    \caption{Sample images from \texttt{RSSC} with regression targets.}
    \label{fig:cloudL9}
\end{figure}

% \subsubsection{Real scenes with synthetic clouds (\texttt{RSSC})}
\noindent\textbf{Real scenes with synthetic clouds (\texttt{RSSC})}. We sourced images from NASA's Landsat-9 Level-1 Collection. Ten scenes from different locations and diverse terrains (\emph{e.g.}, sea, mountain, urban, rural, dessert) were selected. We ensured that the collected scenes had no visible clouds. The 12-band raw data of each scene was converted to top-of-atmosphere spectral radiance, corrected for the sun angle using the metadata, and bands 4-3-2 were then selected to generate an RGB base image of dimension of $8000$x$8000$ pixels.
Synthetic clouds with groundtruth cloud masks were generated and added to the base images using SatelliteCloudGenerator \cite{rs15174138}; see Fig.~\ref{fig:cloudL9}. Then, we cropped the base images and their ground-truth masks into $128$x$128$ non-overlapping patches ($x$). In total, we extracted $9,600$ patches from $5$ base images for EL and $8,000$ patches from the remaining $5$ base images as testing data from unseen before scenes. Using synthetic clouds allowed us to control the groundtruth cloud coverage percentage ($y$) in each image and its distribution over the dataset. \texttt{RSSC} was modified into four variations, each corresponding to a different ground truth label distribution.
\begin{itemize}
    \item\texttt{RSSC-B}: Bimodal distribution,
    \item\texttt{RSSC-N}: Negatively skewed distribution,
    \item\texttt{RSSC-U}: Uniform distribution,
    \item\texttt{RSSC-G}: Gaussian distribution.
\end{itemize}

\begin{figure*}[ht]
    \centering
    \includegraphics[width=1\textwidth]{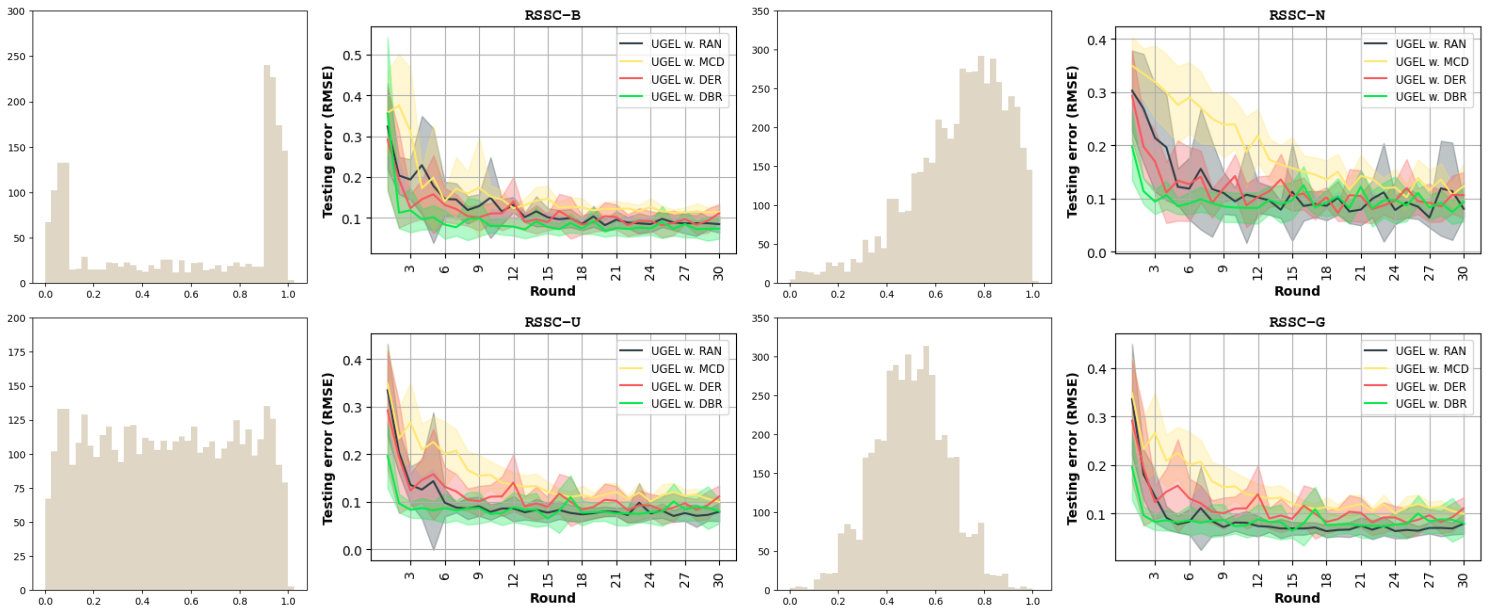}
    \caption{Comparison of different uncertainty estimation methods in UGEL on synthetic datasets: (Top Left) \texttt{RSSC-B}, (Top Right) \texttt{RSSC-N}, (Bottom Left) \texttt{RSSC-U}, (Bottom Right) \texttt{RSSC-G}. Bolded curves and shaded regions respectively indicate mean RMSE testing error and std dev over 10 runs.}
    \label{fig:distribution_ablation}
\end{figure*}

Fig.~\ref{fig:distribution_ablation} shows the performance of different uncertainty estimation methods in UGEL under different ground truth label distributions. Tab.~\ref{tab:pvaluesbenchmark_synthetic} provides Wilcoxon signed-rank test p-values comparing UGEL with DBR to other methods on RSSC using the ResNet18 backbone.

\subsection{Runtime comparisons}

% RSRC-L8 dataset
\begin{table}[ht]
   \small
   \centering
   \begin{tabular}{lccr}
   \toprule\toprule
   \textbf{Method} & \textbf{Uncertain.~est.} & \textbf{Model retrain.} \\ 
   \midrule
            UGEL w.~RAN & n/a & 8.26-12.54\\
            UGEL w.~MCD & 122.19-127.94 & 5.93-12.35\\
            UGEL w.~DER & 18.39-19.04 & 6.06-10.61\\
            UGEL w.~DBR & 18.01-18.63 & 5.96-12.16\\
            
   \bottomrule
   \end{tabular}
   \caption{Runtime (minimum and maximum, in seconds) for different uncertainty estimation methods during the first 10 rounds of UGEL on the \texttt{RSRC-L8} dataset using the ResNet18 backbone.}
   \label{tab:runtime_rsrc_l8}
\end{table}

% RSLC dataset
\begin{table}[ht]
   \small
   \centering
   \begin{tabular}{lccr}
   \toprule\toprule
   \textbf{Method} & \textbf{Uncertain.~est.} & \textbf{Model retrain.} \\ 
   \midrule
            UGEL w.~RAN & n/a & 7.73-9.61\\
            UGEL w.~MCD &  1021.00-1095.18 & 7.83-10.24\\
            UGEL w.~DER & 176.55-507.84 & 7.84-9.65\\
            UGEL w.~DBR & 171.94-507.43 & 7.93-9.32\\
            
   \bottomrule
   \end{tabular}
   \caption{Runtime (minimum and maximum, in seconds) for different uncertainty estimation methods during the first 5 rounds of UGEL on the \texttt{RSLC} dataset  using the ResNet18 backbone.}
   \label{tab:runtime_rslc}
\end{table}

% MobileNetv4
\begin{table}[ht]
   \small
   \centering
   \begin{tabular}{l||lccccr}
   \toprule\toprule
   \textbf{} & \textbf{Test} & \textbf{Rd. 2} & \textbf{Rd. 4} & \textbf{Rd. 6} & \textbf{Rd. 8} \\ 
   \midrule
        \multirow{4}{*}{\rotatebox[origin=c]{90}{\scriptsize\texttt{RSRC-S2}}} & UGEL vs Rand & 0.031 & 0.031 & 0.422 & 0.156 \\
                                                                                & UGEL vs BALD &  0.016 & 0.016 & 0.016 & 0.016 \\
                                                                                & UGEL vs SSL & 0.891 & 0.500 & 0.578 & 0.422 \\
                                                                                & UGEL vs ASSL & 0.016 & 0.047 & 0.078 & 0.016 \\
    \toprule\
     \textbf{} & \textbf{Test} & \textbf{Rd. 2} & \textbf{Rd. 4} & \textbf{Rd. 6} & \textbf{Rd. 8} \\ 
        \midrule
        \multirow{4}{*}{\rotatebox[origin=c]{90}{\scriptsize\texttt{RSRC-L8}}} & UGEL vs Rand & 0.007 & 0.080 & 0.014 & 0.065 \\
                                                                                & UGEL vs BALD & 0.001 & 0.002 & 0.003 & 0.001 \\
                                                                                & UGEL vs SSL & 0.097 & 0.053 & 0.001 & 0.002 \\
                                                                                & UGEL vs ASSL & 0.001 & 0.001 & 0.001 & 0.001 \\
    \toprule\
     \textbf{} & \textbf{Test} & \textbf{Rd. 1} & \textbf{Rd. 2} & \textbf{Rd. 3} & \textbf{Rd. 4} \\ 
        \midrule                                                           
        \multirow{4}{*}{\rotatebox[origin=c]{90}{\scriptsize\texttt{RSLC}}} & UGEL vs Rand & 0.031 & 0.078 & 0.031 & 0.031 \\
                                                                                & UGEL vs BALD & 0.016 & 0.031 & 0.016 & 0.016 \\
                                                                                & UGEL vs SSL &  0.953 & 0.781 &  0.156 & 0.047 \\
                                                                                & UGEL vs ASSL & 0.219 & 0.219 & 0.047 & 0.047 \\
   \bottomrule
   \end{tabular}
   \caption{Wilcoxon signed-rank test p-values comparing UGEL with DBR against other methods on all datasets using the MobileNetV4 backbone.}
   \label{tab:pvaluesbenchmark_mobilenetv4}
\end{table}

\begin{table}[H]
   \small
   \centering
   \begin{tabular}{l||lccccr}
   \toprule\toprule
   
   \textbf{} & \textbf{Test} & \textbf{Rd. 5} & \textbf{Rd. 10} & \textbf{Rd. 15} & \textbf{Rd. 20}  \\  
    \midrule                                                           
        \multirow{3}{*}{\rotatebox[origin=c]{90}{\scriptsize\texttt{RSSC-B}}} & DBR vs RAN & 0.032 & 0.002 & 0.007 & 0.019 \\
                                                                    & DBR vs MCD & 0.002 & 0.001 & 0.001 & 0.001 \\
                                                                    & DBR vs DER  & 0.032 & 0.014 & 0.116 & 0.002 \\
    \toprule
     \textbf{} & \textbf{Test} & \textbf{Rd. 5} & \textbf{Rd. 10} & \textbf{Rd. 15} & \textbf{Rd. 20} \\ 
        \midrule
        \multirow{3}{*}{\rotatebox[origin=c]{90}{\scriptsize\texttt{RSSC-N}}} & DBR vs RAN & 0.01 & 0.188  & 0.461 & 0.539 \\
                                                                    & DBR vs MCD & 0.001 & 0.001 & 0.005 & 0.024 \\
                                                                    & DBR vs DER  & 0.007 & 0.001 & 0.348 & 0.116 \\
    \toprule
     \textbf{} & \textbf{Test} & \textbf{Rd. 5} & \textbf{Rd. 10} & \textbf{Rd. 15} & \textbf{Rd. 20} \\ 
        \midrule                                                           
        \multirow{3}{*}{\rotatebox[origin=c]{90}{\scriptsize\texttt{RSSC-U}}} & DBR vs RAN & 0.08 & 0.348 & 0.116 & 0.423 \\
                                                                    & DBR vs MCD & 0.001 & 0.001 & 0.001 & 0.003 \\
                                                                    & DBR vs DER  & 0.005 & 0.019 & 0.014 & 0.024 \\
    \toprule
     \textbf{} & \textbf{Test} & \textbf{Rd. 5} & \textbf{Rd. 10} & \textbf{Rd. 15} & \textbf{Rd. 20} \\ 
        \midrule
        \multirow{3}{*}{\rotatebox[origin=c]{90}{\scriptsize\texttt{RSSC-G}}} & DBR vs RAN & 0.539 & 0.348 & 0.278 & 0.935 \\
                                                                    & DBR vs MCD & 0.001 & 0.001 & 0.001 & 0.003 \\
                                                                    & DBR vs DER  & 0.005 & 0.019 & 0.014 & 0.024 \\
    
   \bottomrule
   \end{tabular}
   \caption{Wilcoxon signed-rank test p-values comparing UGEL with DBR to other methods on \texttt{RSSC} using the ResNet18 backbone.}
   \label{tab:pvaluesbenchmark_synthetic}
\end{table}

% We can compute the negative log likelihood loss, $\cL_{NLL}$, as:
% \begin{equation}
% \begin{split}
%   \cL_{NLL}(\mu. \nu, y)  = -[\log{\Gamma(\nu)} - \log{\Gamma(\mu\nu)}
%   - \log{\Gamma((1-\mu)\nu)}\\ + \log{y^{\mu\nu-1}} + \log{(1-y)^{(1-\mu)\nu-1}}]\\
%   = -\log{\Gamma(\nu)} + \log{\Gamma(\mu\nu)} + \log{\Gamma((1-\mu)\nu)}\\
%   - (\mu\nu-1)\log{y} + ((1-\mu)\nu-1)\log{(1-y)}
% \end{split}
% \end{equation}\label{eq:likelihood}

\end{document}